%% file: tea_arxiv_2023.tex
\pdfoutput=1

\documentclass[11pt]{article}

\usepackage{ACL2023}
\usepackage{times}
\usepackage{latexsym}

\usepackage[T1]{fontenc}

\usepackage[utf8]{inputenc}

\usepackage{microtype}

\usepackage{inconsolata}

\usepackage{algorithm}%
\usepackage[noend]{algpseudocode}%
\usepackage{tikz}%
\usepackage{pgfplots}%
\usepackage{array}
\usepackage{pgfplotstable}%
\pgfplotsset{compat=newest}%
\usepgfplotslibrary{fillbetween}%
\usepackage{multirow}%
\usepackage{pdfpages}%
\usepackage{longtable}%
\input{utils/math_commands.tex}%
\input{utils/commands.tex}%
\input{utils/constants.tex}%
\input{utils/colors.tex}%
\input{utils/line_styles.tex}%
\input{utils/plot_commands.tex}
\title{Estimating the Adversarial Robustness of Attributions in Text with Transformers}%
 \author{%
 	Adam Ivankay, Mattia Rigotti, Ivan Girardi \and Chiara Marchiori\\%
 	IBM Research Zurich\\%
 	R\"{u}schlikon, Switzerland\\%
 	\texttt{\{aiv,mrg,ivg,chi\}@zurich.ibm.com}%
	\AND%
	Pascal Frossard\\%
	\'{E}cole Polytechnique F\'{e}d\'{e}rale de Lausanne (EPFL)\\%
	Lausanne, Switzerland\\%
	\texttt{pascal.frossard@epfl.ch}%
}%
\begin{document}%
\maketitle%
%
\input{src/abstract.tex}

\input{src/intro.tex}

\input{src/related_work.tex}

\input{src/prelim.tex}

\input{src/methods.tex}

\input{src/results.tex}

\input{src/conclusion.tex}


\bibliographystyle{acl_natbib}
\bibliography{bibliography}

\newpage
\appendix
\setcounter{page}{1}


\section*{Supplementary material}

# A Appendix

## A.1 Datasets

We estimate the robustness of our attribution methods and models on five publicly available datasets. These are AG's News, MR movie review, IMDB movie review, Yelp and Fake News, all of which are English language. AG's News consists of 127552 news article samples, categorized into the classes World, News, Business and Sci/Tech. We use the concatenation of title and text of the samples to feed into our text classifiers, stripping any sample that is longer than 64 tokens. The MR movie review dataset contains 10592 short samples of positive or negative movie reviews. We only use the first 32 tokens in each sample as input to the classifiers. IMDB movie review is a dataset consisting of 49952 positive and negative movie reviews, with a maximum token length of 256. Yelp categorizes 700000 reviews of several topics into 5 classes, each representing a rating from 1 to 5. We strip the samples to a maximum length of 256. Fake News is a collection of 20080 news samples, each categorized into reliable or unreliable. These are rather long articles, thus we use a maximum sequence length of 512 for this dataset.

We apply basic preprocessing to all samples in each dataset, which includes converting them to lowercase, removing any special characters not in the English alphabet and emojis. We use 60% of the samples for training the classifier models, 20% for validation and 20% for testing and estimating the robustness of attribution methods.

| DATASET | CNN | LSTM | LSTMATT | BERT | RoBERTa | XLNET |
|---|---|---|---|---|---|---|
| AG'S NEWS | 89.7% | 90.8% | 91.4% | 94.2% | 94.0% | 93.8% |
| MR | 73.0% | 76.4% | 78.0% | 82.2% | 87.7% | 86.3% |
| IMDB | 82.0% | 87.2% | 87.3% | 89.4% | 93.3% | 93.7% |
| YELP | 49.0% | 54.8% | 60.0% | 62.6% | 67.6% | - |
| FAKE NEWS | 98.9% | 99.6% | 99.6% | 99.8% | 100.0% | 100.0% |

Table 1: Classifier accuracies

## A.2 Models

| | | AG's News | MR | IMDB | Yelp | Fake News |
|---|---|---|---|---|---|---|
| **CNN** | Input shape | (64, 300) | (32, 300) | (256, 300) | (256, 300) | (512, 300) |
| | Num. classes | 4 | 2 | 2 | 5 | 2 |
| | Filter sizes | [3, 5, 7] | [3, 5] | [3, 5, 7] | [3, 5, 7] | [3, 5, 7] |
| | Feature sizes | [8, 8, 8] | [8, 8] | [16, 16, 16] | [128, 128, 128] | [32, 32, 32] |
| | Pooling sizes | [2, 2, 2] | [2, 2] | [2, 2, 2] | [2, 2, 2] | [2, 2, 2] |
| | Lin. layer dim. | 8 | 8 | 16 | 64 | 32 |
| | Num. params | 67748 | 27946 | 567458 | 16428293 | 4091714 |
| **LSTM** | Input shape | (64, 300) | (32, 300) | (256, 300) | (256, 300) | (512, 300) |
| | Num. classes | 4 | 2 | 2 | 5 | 2 |
| | Hidden dim. | 8 | 8 | 16 | 256 | 16 |
| | Num. layers | 1 | 1 | 2 | 2 | 1 |
| | Pooling sizes | 2 | 2 | 1 | 2 | 2 |
| | Lin. layer dim. | 8 | 8 | 16 | 32 | 16 |
| | Num. params | 10988 | 10458 | 18162 | 2146693 | 85986 |
| **LSTMAtt** | Input shape | (64, 300) | (32, 300) | (256, 300) | (256, 300) | (512, 300) |
| | Num. classes | 4 | 2 | 2 | 5 | 2 |
| | Hidden dim. | 8 | 8 | 16 | 256 | 16 |
| | Num. layers | 4 | 1 | 2 | 2 | 1 |
| | Lin. layer dim. | 8 | 8 | 16 | 32 | 16 |
| | Num. params | 25004 | 19994 | 47666 | 2752901 | 41826 |
| **BERT** | Input shape | (64,) | (32,) | (256,) | (256,) | (512,) |
| | Num. classes | 4 | 2 | 2 | 5 | 2 |
| | Model ID | bert-base-uncased | bert-base-uncased | bert-base-uncased | bert-base-uncased | bert-base-uncased |
| | Num. params | 109485316 | 109483778 | 109483778 | 109486085 | 109483778 |
| **RoBERTa** | Input shape | (64,) | (32,) | (256,) | (256,) | (512,) |
| | Num. classes | 4 | 2 | 2 | 5 | 2 |
| | Model ID | roberta-base | roberta-base | roberta-base | roberta-base | roberta-base |
| | Num. params | 124648708 | 124647170 | 124647170 | 124649477 | 124647170 |
| **XLNet** | Input shape | (64,) | (32,) | (256,) | (256,) | (512,) |
| | Num. classes | 4 | 2 | 2 | 5 | 2 |
| | Model ID | xlnet-base-cased | xlnet-base-cased | xlnet-base-cased | xlnet-base-cased | xlnet-base-cased |
| | Num. params | 117312004 | 117310466 | 117310466 | 117312773 | 117310466 |

Table 2: Model specifications

As described in the main paper, we train six classification architectures for each dataset, three DNN-based architectures, which are a CNN, an LSTM, an LSTM with an attention layer (LSTMAtt), as well as three transformer-based architectures, which are a finetuned BERT, RoBERTa and XLNet. The CNN, LSTM and LSTMAtt architectures use the 6B-300-dimensional Glove word embeddings, while the transformer-based architectures use the pretrained Huggingface embeddings of the respective base-uncased versions. During training, the embeddings are not trained, only the model parameters. The DNN-based classifiers each contain a linear layer on top of their feature extractors and use the built-in SpaCy English tokenizer, the transformers directly map the feature outputs to the output logits with a fully-connected layer and utilize the Huggingface preptrained tokenizers for each architecture respectively. Table 2 contains the model specifications. We train each model with a standard learning rate of 0.001, using the cross-entropy loss and early stopping. We utilize NVIDIA A100 GPUs to speed up training and AR estimation. The resulting accuracies of the models can be found in Table 1.

### A.3 Additional AR results

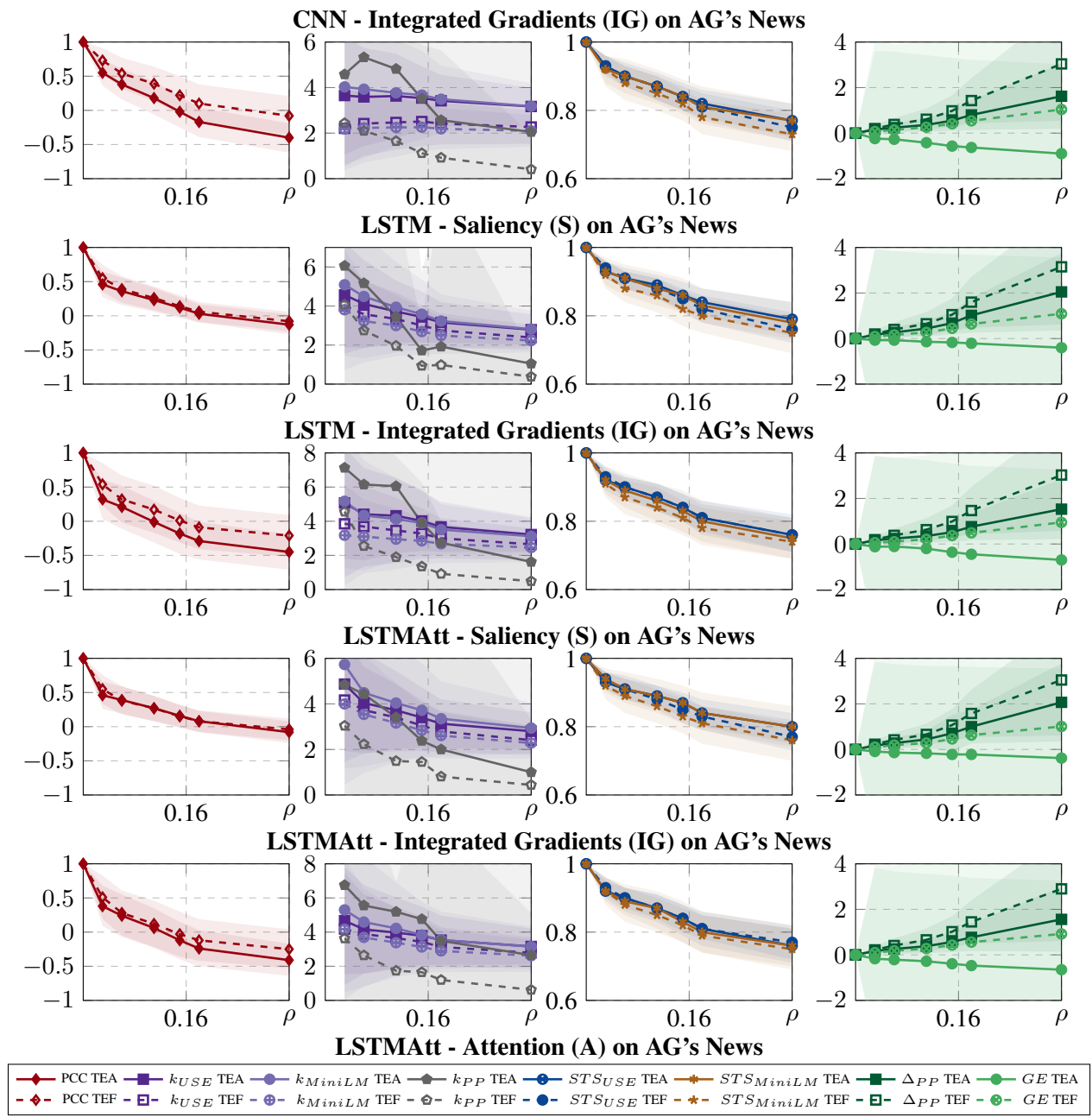

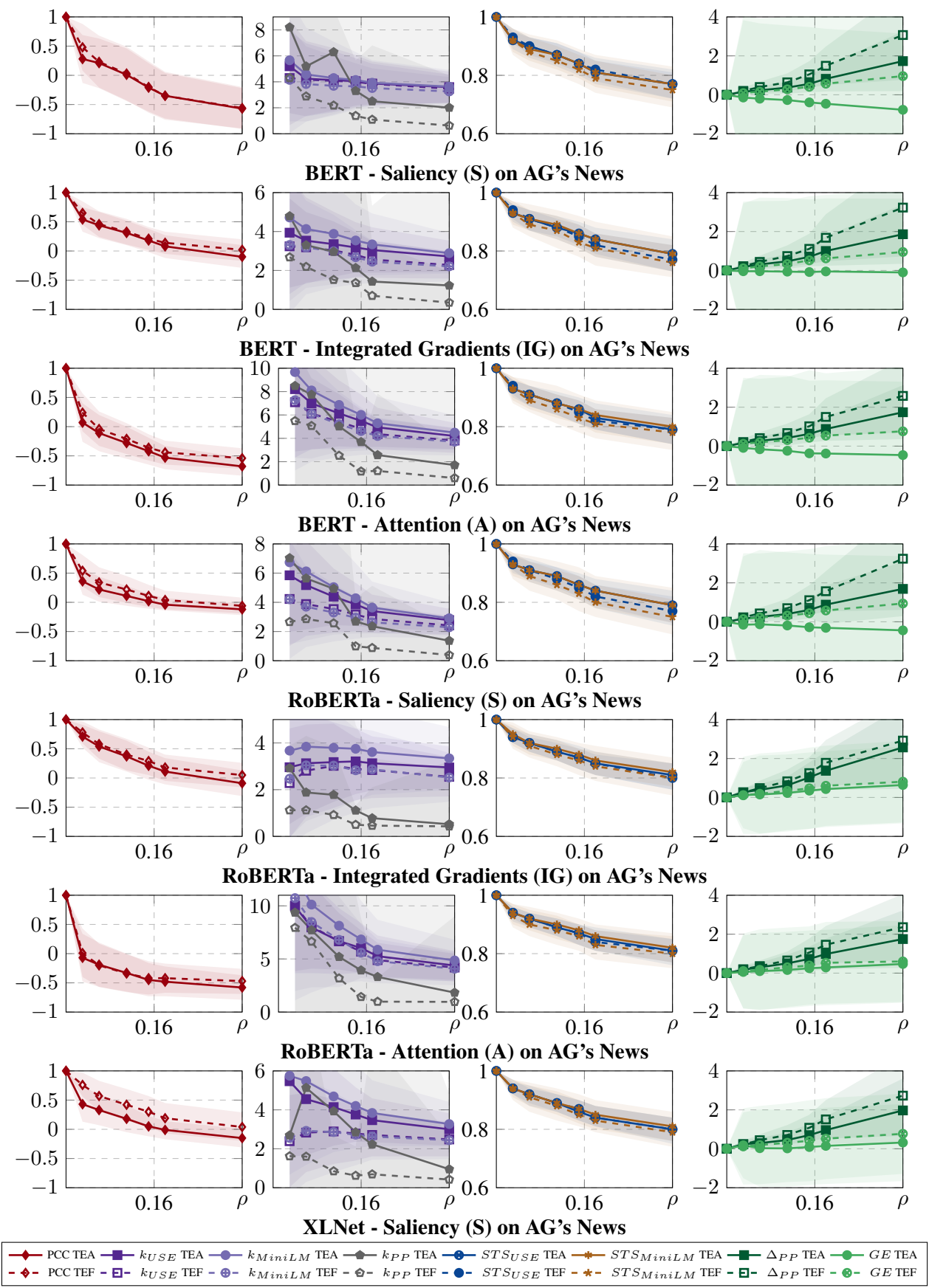

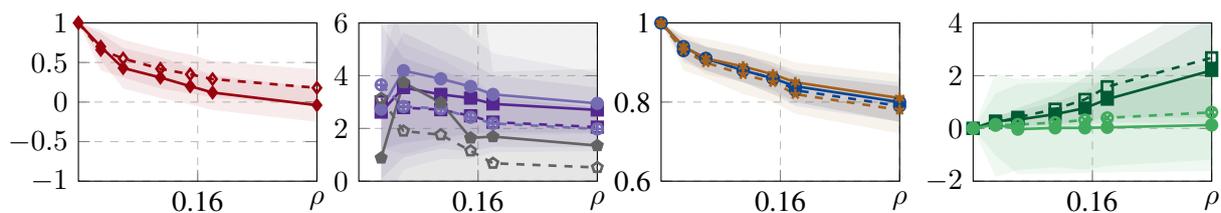
XLNet - Integrated Gradients (IG) on AG's News

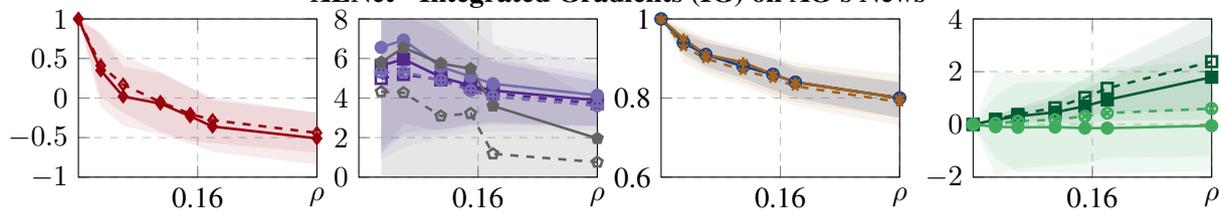
XLNet - Attention (A) on AG's News

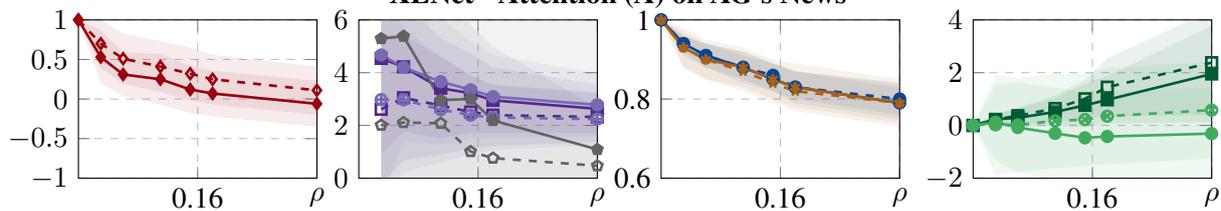
CNN - Saliency (S) on MR

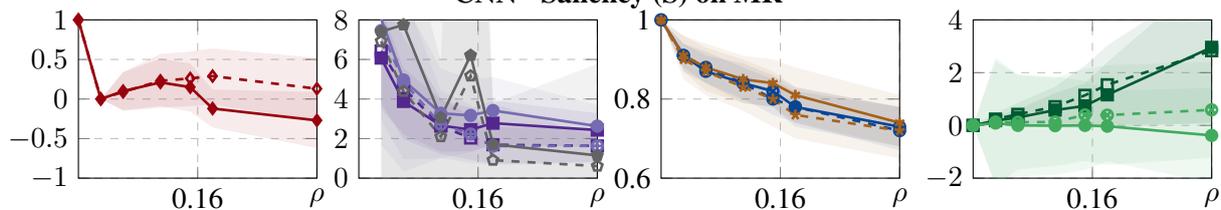
CNN - Integrated Gradients (IG) on MR

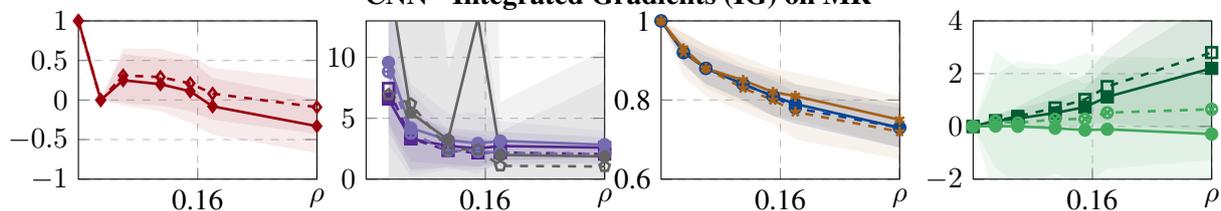
LSTM - Saliency (S) on MR

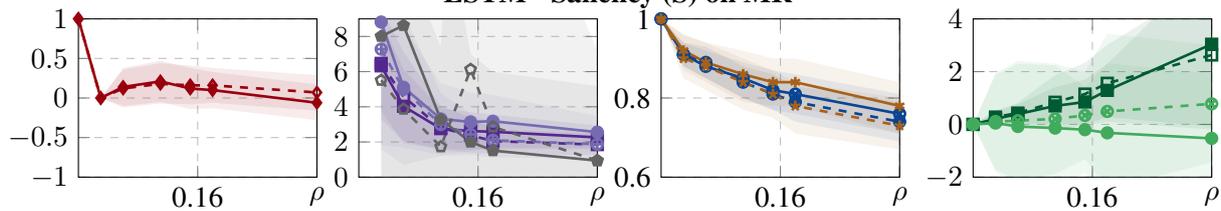
LSTM - Integrated Gradients (IG) on MR

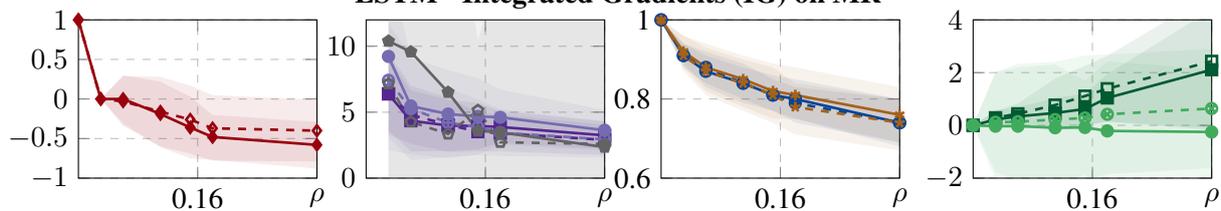
LSTMAtt - Saliency (S) on MR

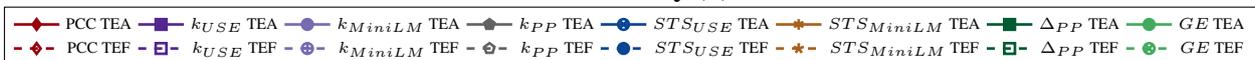

4

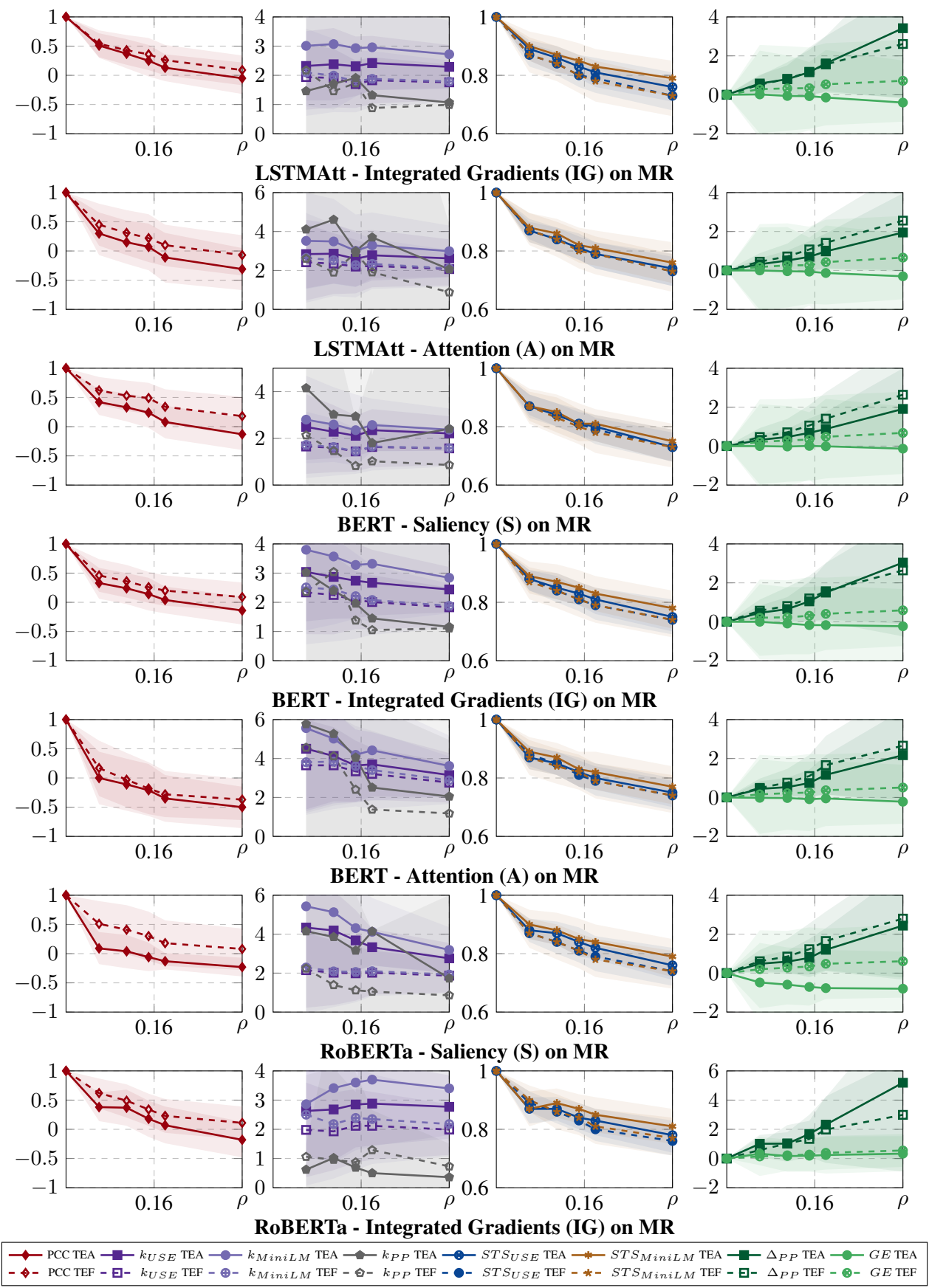

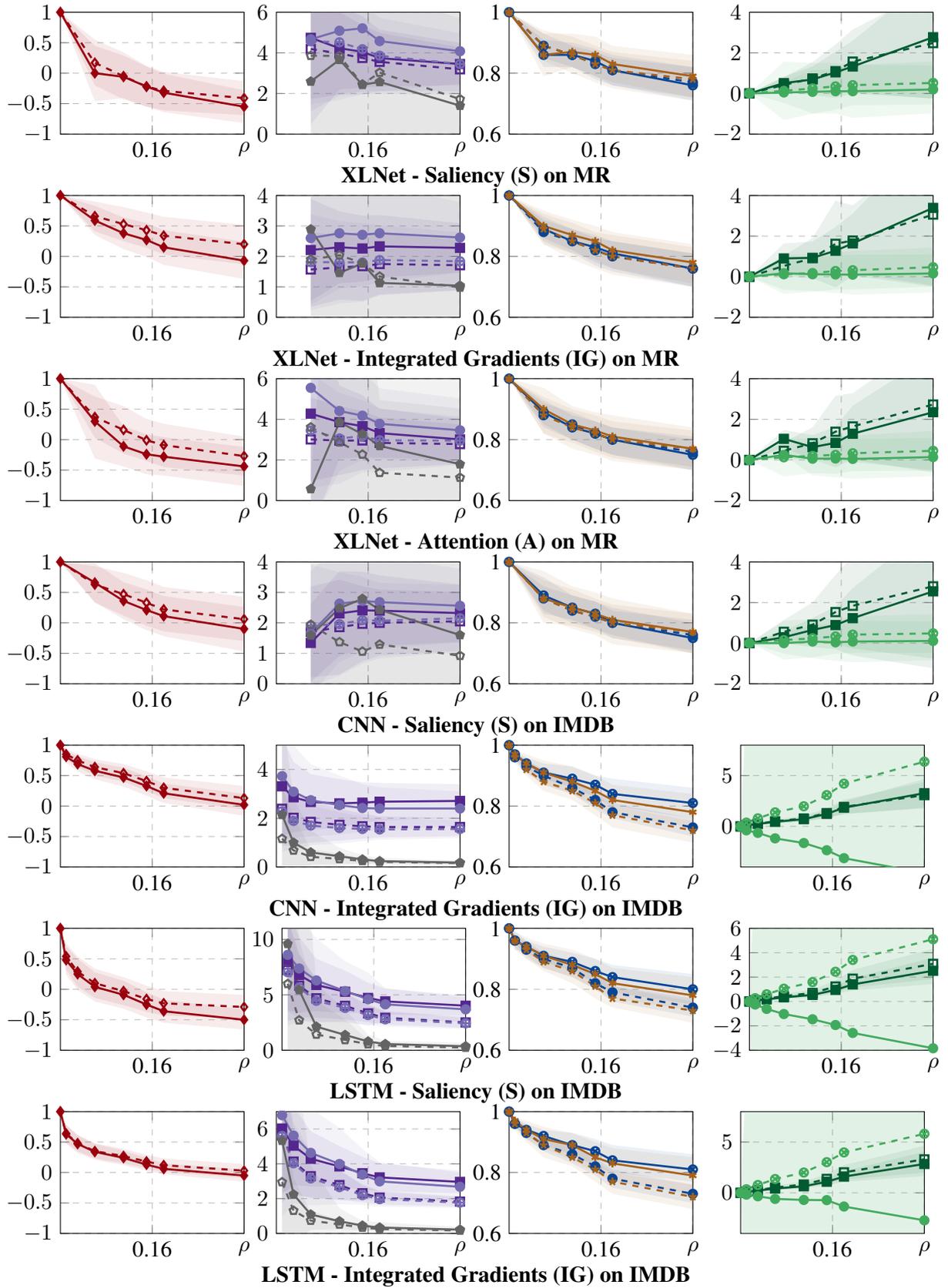

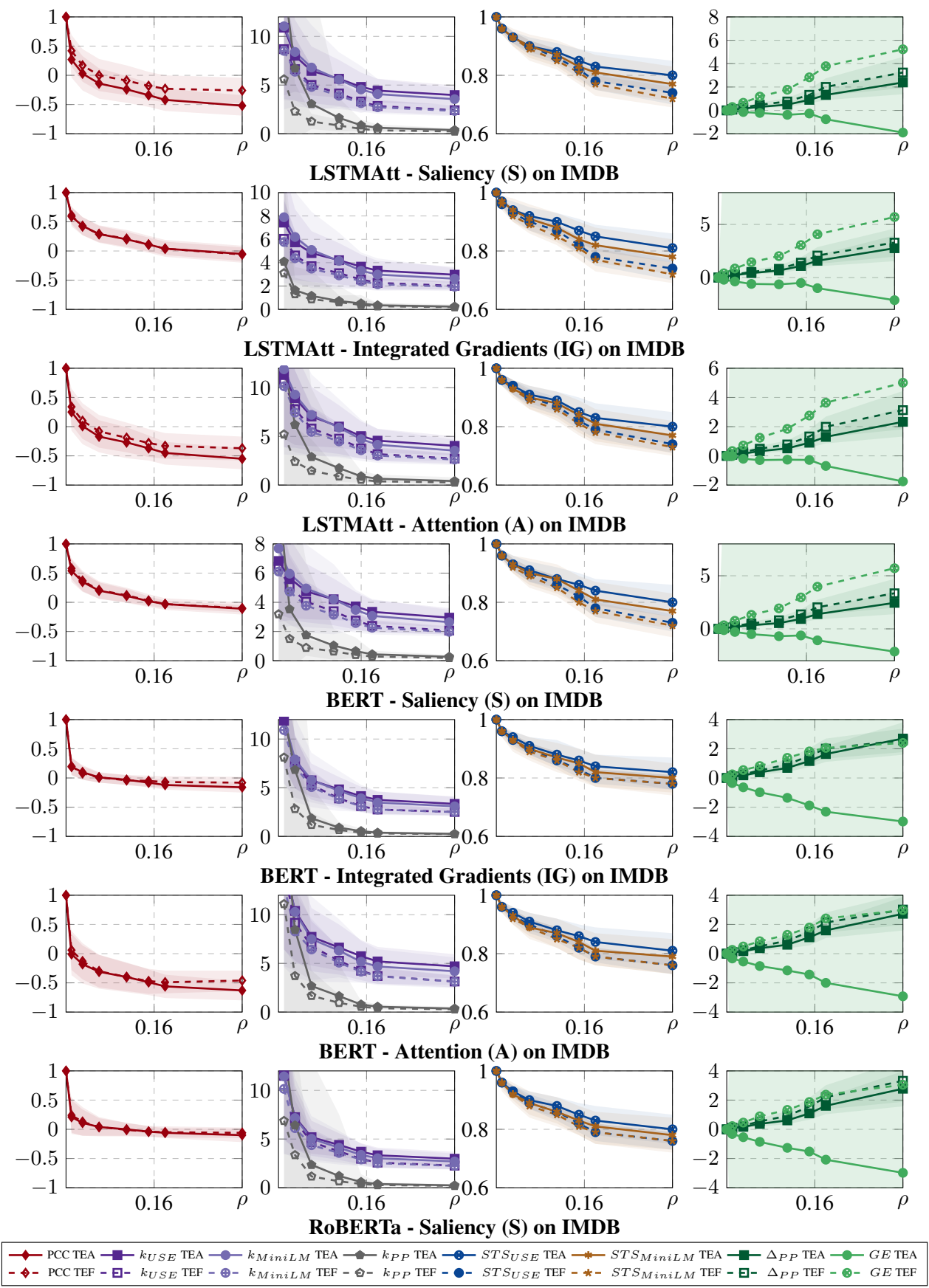

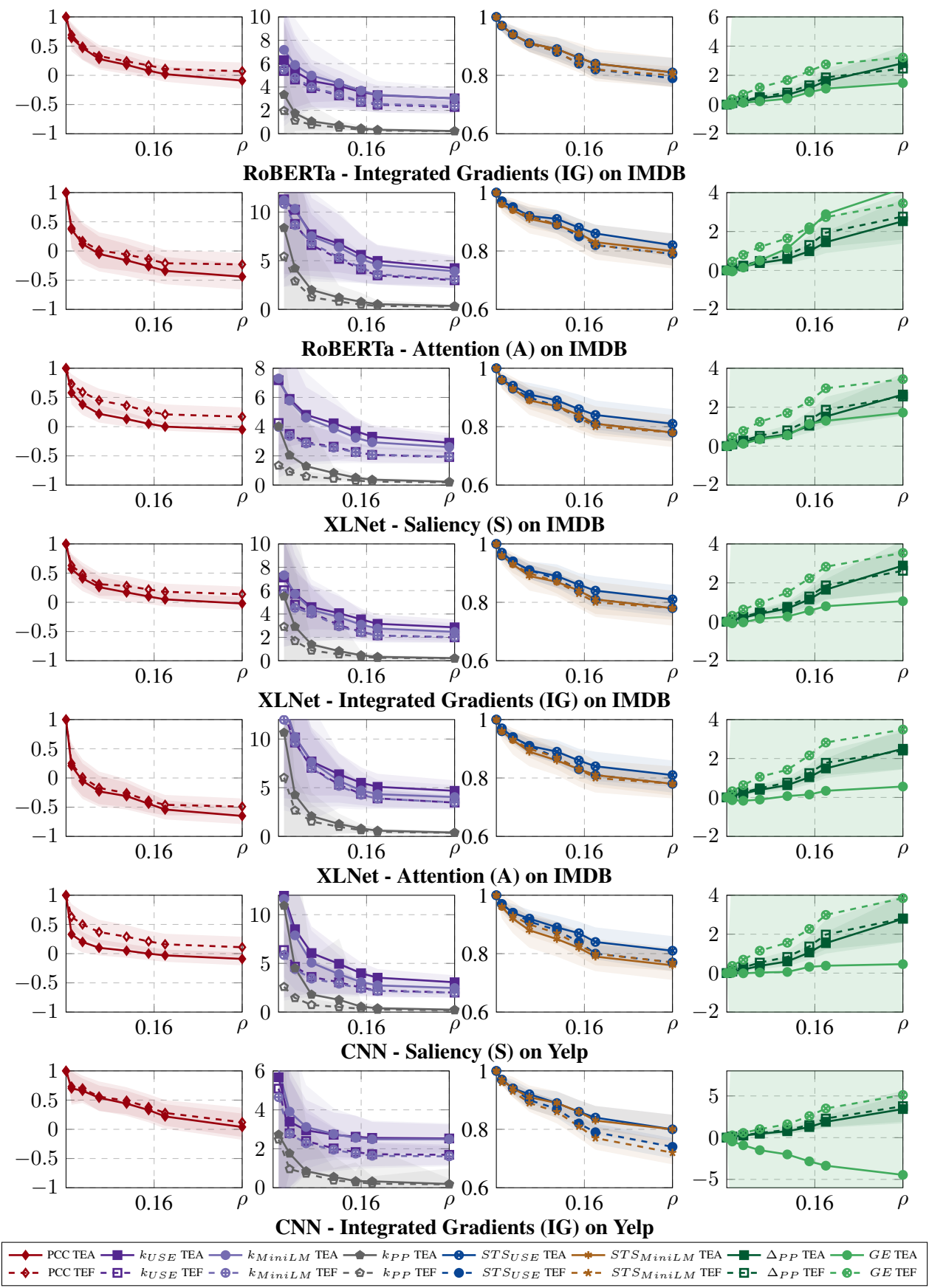

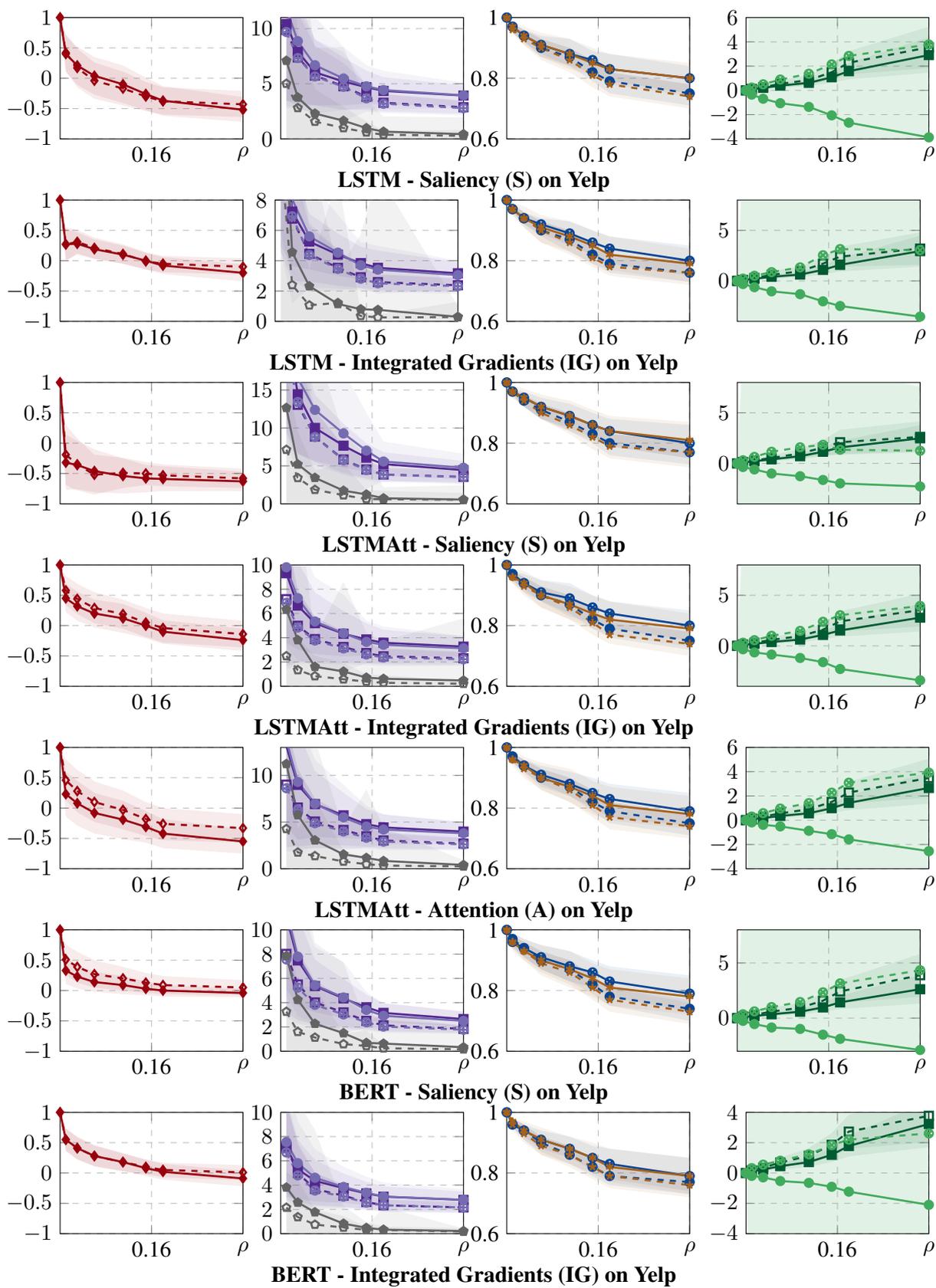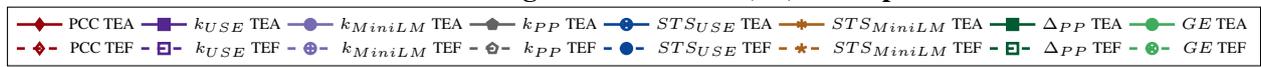

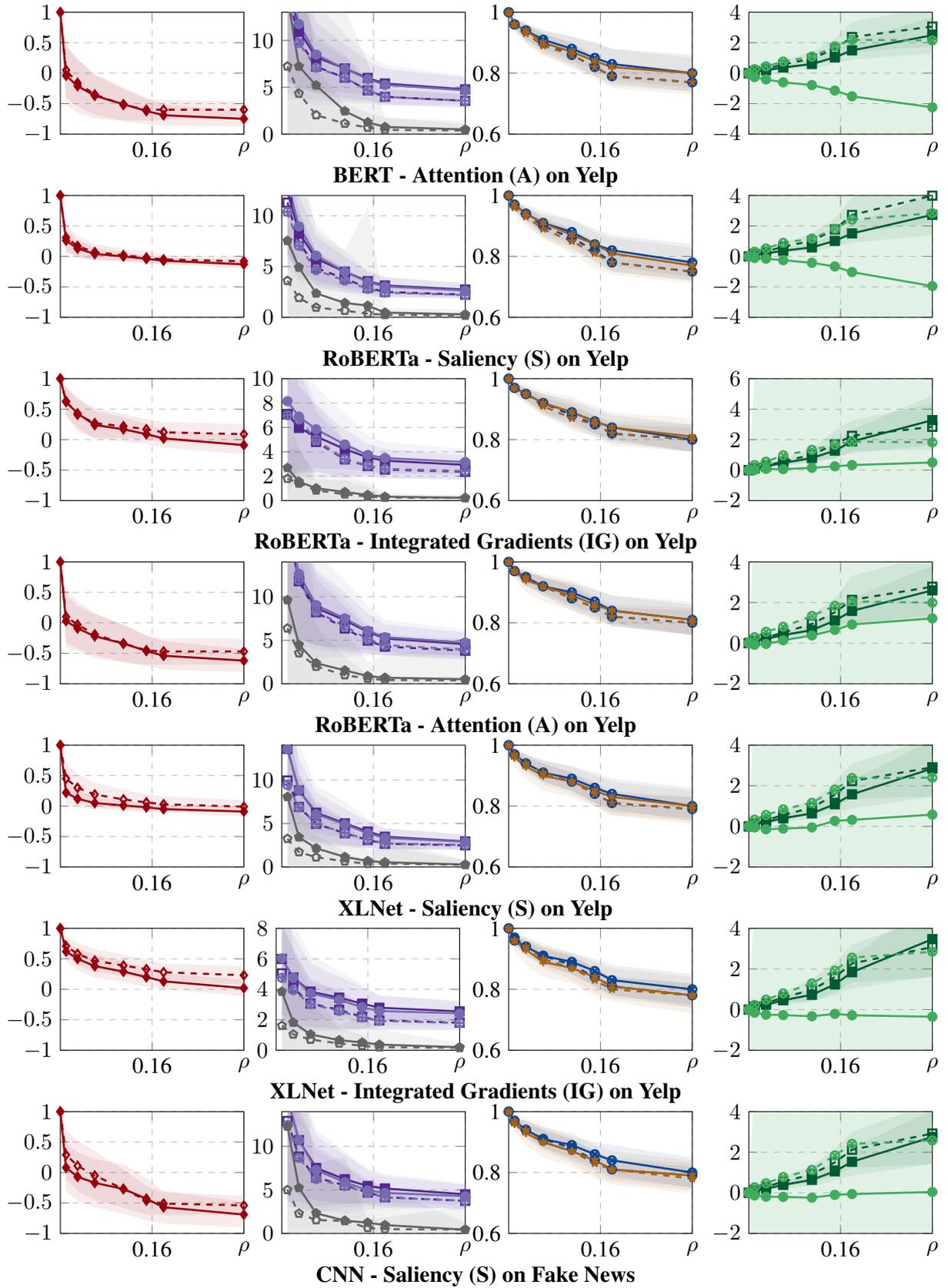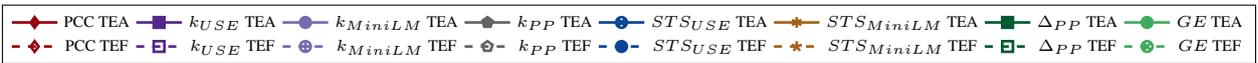

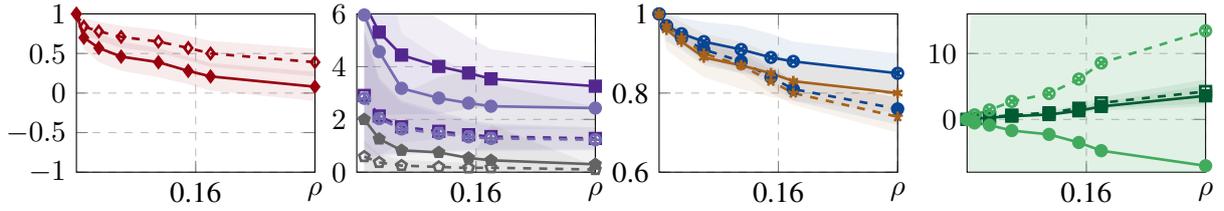
**CNN - Integrated Gradients (IG) on Fake News**

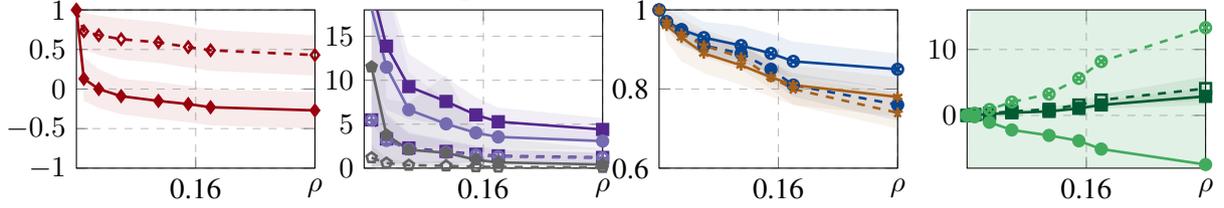
**LSTM - Saliency (S) on Fake News**

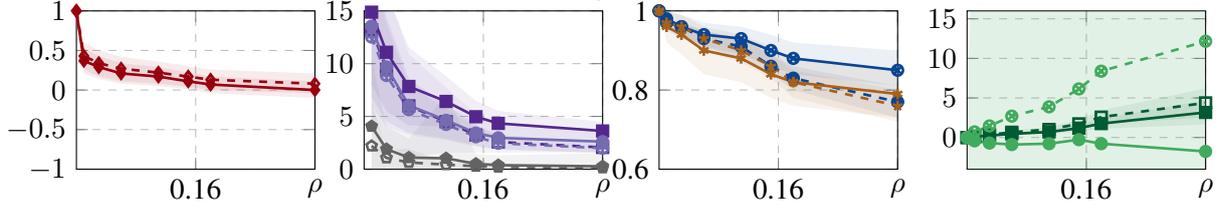
**LSTM - Integrated Gradients (IG) on Fake News**

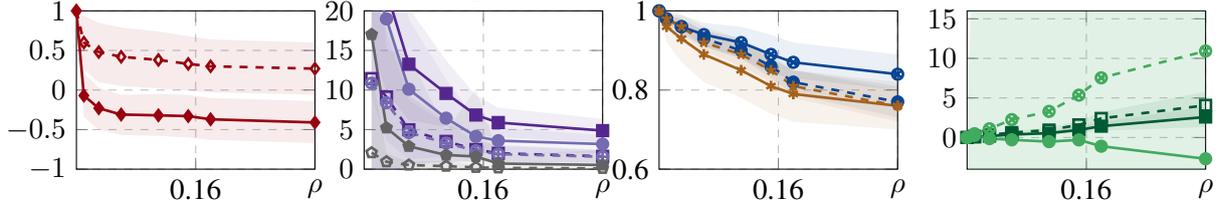
**LSTMAtt - Saliency (S) on Fake News**

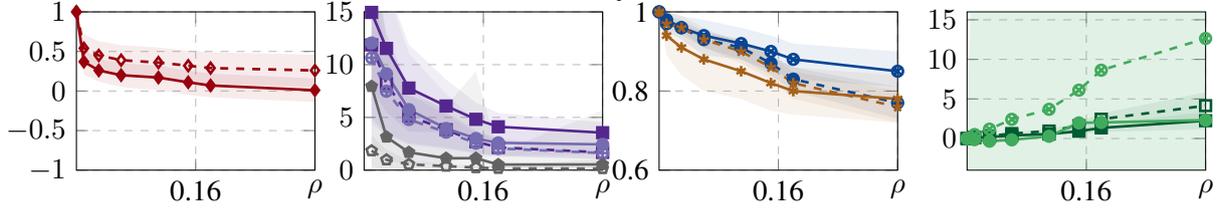
**LSTMAtt - Attention (A) on Fake News**

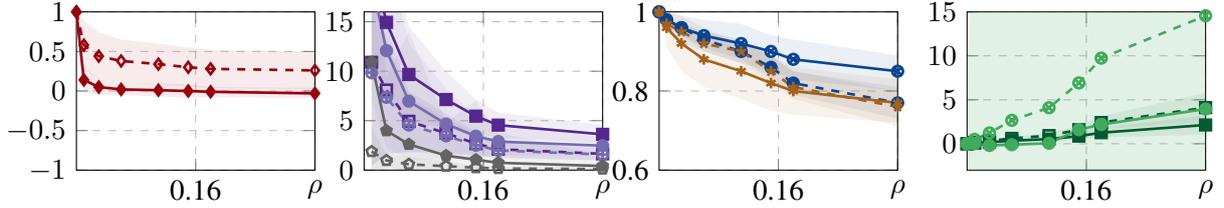
**BERT - Saliency (S) on Fake News**

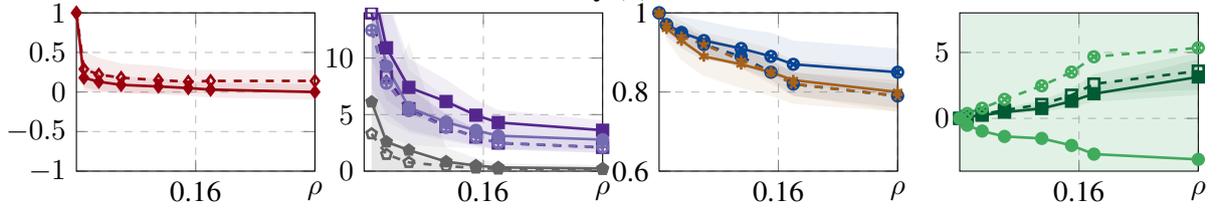
**BERT - Integrated Gradients (IG) on Fake News**

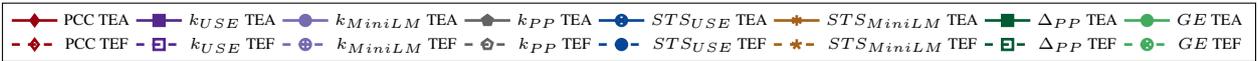

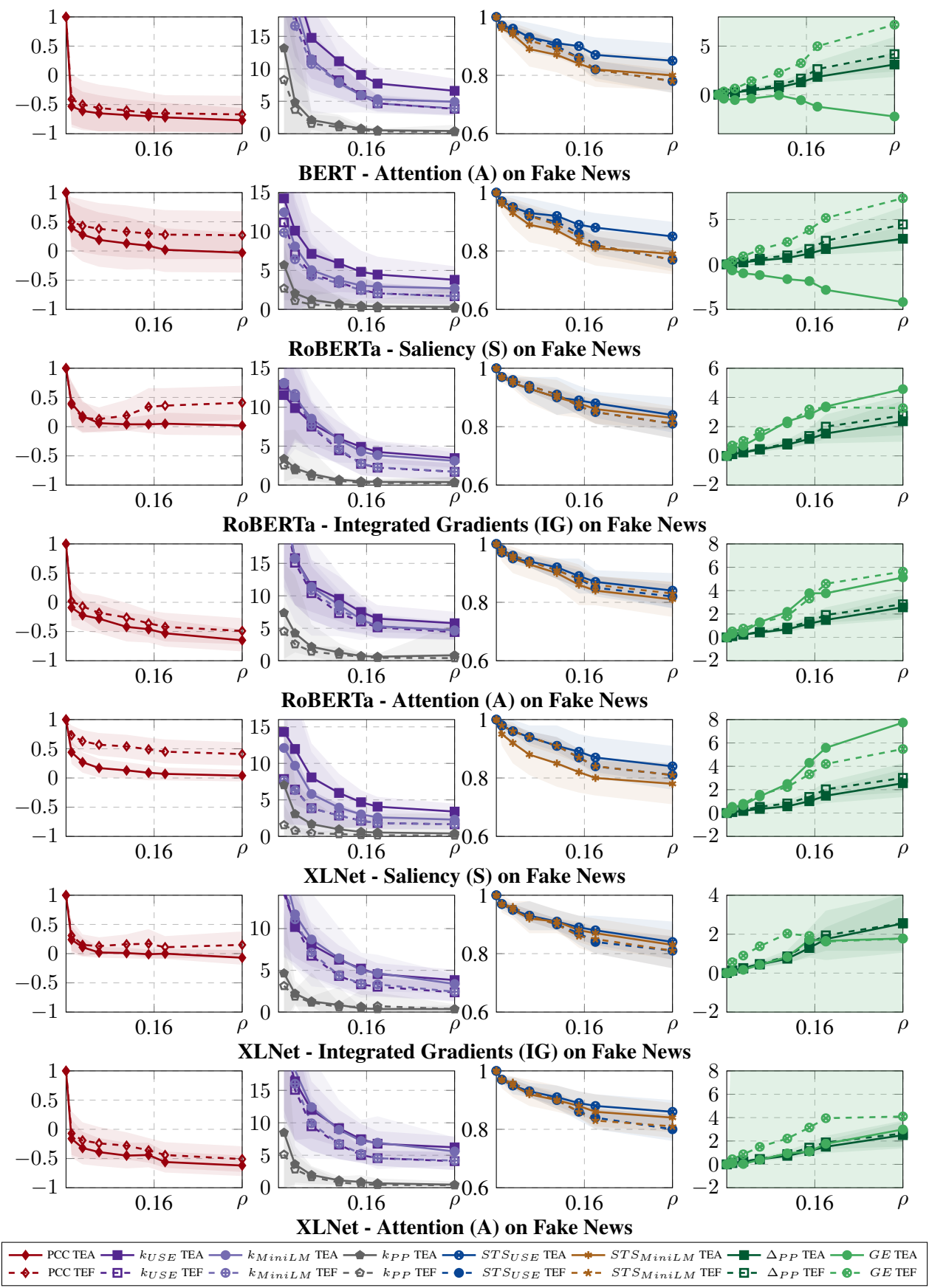

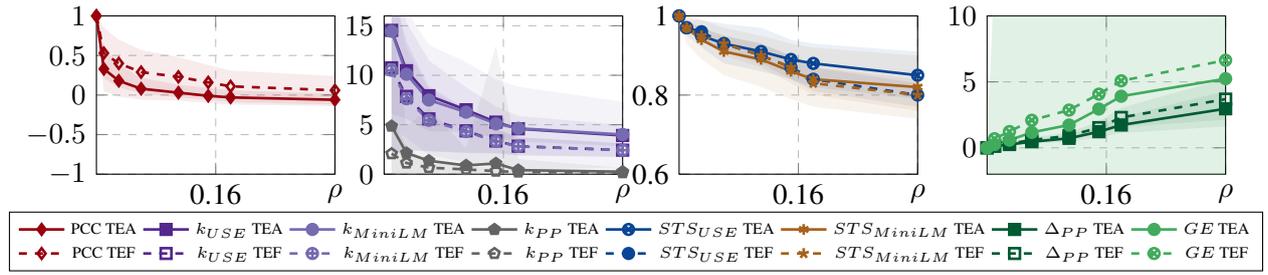




\end{document}

%% file: utils/math_commands.tex
\usepackage{amsmath,amsfonts,bm}

\def\eqref#1{equation~\ref{#1}}

\def\1{\bm{1}}

\def\va{{\bm{a}}}

\def\vm{{\bm{m}}}

\def\vo{{\bm{o}}}

\def\vs{{\bm{s}}}

\def\evc{{c}}

\def\evl{{l}}

\def\evw{{w}}

\def\mX{{\bm{X}}}

\DeclareMathAlphabet{\mathsfit}{\encodingdefault}{\sfdefault}{m}{sl}
\SetMathAlphabet{\mathsfit}{bold}{\encodingdefault}{\sfdefault}{bx}{n}

\def\sC{{\mathbb{C}}}

\def\sL{{\mathbb{L}}}

\def\sS{{\mathbb{S}}}

\def\sW{{\mathbb{W}}}

\newcommand{\R}{\mathbb{R}}

\DeclareMathOperator*{\argmax}{arg\,max}

%% file: utils/commands.tex
\usepackage{pgfkeys}

\newcommand{\setval}[1]{\pgfkeys{/variables/#1}}
\newcommand{\getval}[1]{\pgfkeysvalueof{/variables/#1}}
\newcommand{\declare}[1]{%
 \pgfkeys{
  /variables/#1.is family,
  /variables/#1.unknown/.style = {\pgfkeyscurrentpath/\pgfkeyscurrentname/.initial = ##1}
 }%
}

\declare{}

\newcount\nbofwords
\makeatletter  
\def\myutil@empty{}
\def\multiwords#1 #2\@nil{%
 \def\NextArg{#2}%
 \advance\nbofwords by  1 %
 \expandafter\edef\csname word\@alph\nbofwords\endcsname{#1}%
 \ifx\myutil@empty\NextArg
     \let\next\@gobble
 \fi
 \next#2\@nil
}%

\def\GetWords#1{%
   \let\next\multiwords 
   \nbofwords=0 %
   \expandafter\next#1 \@nil %
}%
\makeatother

%% file: utils/constants.tex
\setval{dataroot = data/mlm_plots}

\setval{timingroot = \getval{dataroot}/timing_data}
\setval{tefroot = \getval{dataroot}/tef}
\setval{mlmroot = \getval{dataroot}/mlm}
\setval{mlmfastroot = \getval{dataroot}/mlm_fast}
\setval{mlmdistilroot = \getval{dataroot}/mlm_distil}
\setval{mlmdistilfastroot = \getval{dataroot}/mlm_distil_fast}

\newlength{\plotwidth}
\newlength{\plotheight}

\newlength{\rowsepoffset}

\setval{x_min = 0.0}

\setval{x_max = 0.32}
\setval{x_mid_tick = 0.16} 

\setval{y_min = -1.0}
\setval{y_max = 1.0}
\setval{y_mid_tick = 0.0} 

\setval{colorpalette = r3}

\setval{line_metric = mean}
\setval{fill_metric_high = mms}
\setval{fill_metric_low = mps}

\setval{fillopacity = 0.08}
\setval{lineopacity = 1.0}

\newcommand{\tmpvalidmodelexpls}[1]{\def\validmodelexplainers{#1}}
\tmpvalidmodelexpls{{cnn s},{cnn ig},{lstm s},{lstm ig},{lstmatt s},{lstmatt ig},{lstmatt a},{bert s},{bert ig},{bert a},{roberta s},{roberta ig},{roberta a},{xlnet s},{xlnet ig},{xlnet a}}

\setval{d_agnews = AG's News}
\setval{d_mr = MR}
\setval{d_imdb = IMDB}
\setval{d_yelp = Yelp}
\setval{d_fakenews = FakeNews}

\setval{m_cnn = CNN}
\setval{m_lstm = LSTM}
\setval{m_lstmatt = LSTMAtt}
\setval{m_bert = BERT}
\setval{m_roberta = RoBERTa}
\setval{m_xlnet = XLNet}

\setval{e_s = Saliency (S)}
\setval{e_ig = Integrated Gradients (IG)}
\setval{e_a = Attention (A)}

\setval{mf_pcc = correlations}
\setval{mf_spearman = correlations}
\setval{mf_kendall = correlations}

\setval{mf_sts_use = semantic_sims}
\setval{mf_sts_tse = semantic_sims}
\setval{mf_sts_ce = semantic_sims}

\setval{mf_top10 = top_intersections}
\setval{mf_top30 = top_intersections}
\setval{mf_top50 = top_intersections}

\setval{mf_pp_r_d = pp_distances}
\setval{mf_pp_h_d = pp_distances}
\setval{mf_pp_l2_d = pp_distances}

\setval{mf_orig_pp = comparison_vals}
\setval{mf_adv_pp = comparison_vals}
\setval{mf_orig_ge = comparison_vals}
\setval{mf_adv_ge = comparison_vals}

\setval{mf_et_use = ets_sts}
\setval{mf_et_tse = ets_sts}
\setval{mf_et_ce = ets_sts}

\setval{mf_et_p = ets_time_misc}
\setval{mf_et_p_h = ets_time_misc}
\setval{mf_et_p_l2 = ets_time_misc}

\setval{mf_attack_time = time_misc}
\setval{mf_l2 = time_misc}
\setval{mf_lom = time_misc}

%% file: utils/colors.tex
\definecolor{c0}{RGB}{178,24,43}
\definecolor{c1}{RGB}{239,138,98}
\definecolor{c2}{RGB}{253,219,199}
\definecolor{c3}{RGB}{209,229,240}
\definecolor{c4}{RGB}{103,169,207}
\definecolor{c5}{RGB}{33,102,172}

\definecolor{g76}{RGB}{118,42,131}
\definecolor{g75}{RGB}{175,141,195}
\definecolor{g72}{RGB}{217,240,211}
\definecolor{g71}{RGB}{127,191,123}
\definecolor{g70}{RGB}{27,120,55}

\definecolor{rg54}{RGB}{202,0,32}
\definecolor{rg53}{RGB}{244,165,130}
\definecolor{rg52}{RGB}{255,255,255}
\definecolor{rg51}{RGB}{186,186,186}
\definecolor{rg50}{RGB}{64,64,64}

\definecolor{r32}{RGB}{252,146,114}
\definecolor{r31}{RGB}{239,59,44}
\definecolor{r30}{RGB}{153,0,13}

\definecolor{r43}{RGB}{254,229,217}
\definecolor{r42}{RGB}{252,174,145}
\definecolor{r41}{RGB}{251,106,74}
\definecolor{r40}{RGB}{203,24,29}

\definecolor{b32}{RGB}{166,97,26}
\definecolor{b31}{RGB}{66,146,198}
\definecolor{b30}{RGB}{8,69,148}

\definecolor{b43}{RGB}{239,243,255}
\definecolor{b42}{RGB}{189,215,231}
\definecolor{b41}{RGB}{107,174,214}
\definecolor{b40}{RGB}{33,113,181}

\definecolor{g32}{RGB}{161,217,155}
\definecolor{g31}{RGB}{65,171,93}
\definecolor{g30}{RGB}{0,90,50}

\definecolor{v32}{RGB}{99,99,99}
\definecolor{v31}{RGB}{117,107,177}
\definecolor{v30}{RGB}{84,39,143}

%% file: utils/line_styles.tex
\tikzstyle{pccbeastyle} = [color=r30, mark=diamond*, line width=1pt, opacity=\getval{lineopacity}]
\tikzstyle{pcctefstyle} = [dashed, mark options=solid, color=r30, mark=diamond, line width=1pt, opacity=\getval{lineopacity}]

\tikzstyle{kusebeastyle} = [color=v30, mark=square*, line width=1pt, opacity=\getval{lineopacity}]
\tikzstyle{kusetefstyle} = [dashed, mark options=solid, color=v30, mark=square, line width=1pt, opacity=\getval{lineopacity}]

\tikzstyle{ktsebeastyle} = [color=v31, mark=oplus*, line width=1pt, opacity=\getval{lineopacity}]
\tikzstyle{ktsetefstyle} = [dashed, mark options=solid, color=v31, mark=oplus, line width=1pt, opacity=\getval{lineopacity}]

\tikzstyle{kppbeastyle} = [color=v32, mark=pentagon*, line width=1pt, opacity=\getval{lineopacity}]
\tikzstyle{kpptefstyle} = [dashed, mark options=solid, color=v32, mark=pentagon, line width=1pt, opacity=\getval{lineopacity}]

\tikzstyle{stsusebeastyle} = [color=b30, mark=otimes, line width=1pt, opacity=\getval{lineopacity}]
\tikzstyle{stsusetefstyle} = [dashed, mark options=solid, color=b30, mark=otimes*, line width=1pt, opacity=\getval{lineopacity}]

\tikzstyle{ststsebeastyle} = [color=b32, mark=asterisk, line width=1pt, opacity=\getval{lineopacity}]
\tikzstyle{ststsetefstyle} = [dashed, mark options=solid, color=b32, mark=star, line width=1pt, opacity=\getval{lineopacity}]

\tikzstyle{ppbeastyle} = [color=g30, mark=square*, line width=1pt, opacity=\getval{lineopacity}]
\tikzstyle{pptefstyle} = [dashed, mark options=solid, color=g30, mark=square, line width=1pt, opacity=\getval{lineopacity}]

\tikzstyle{gebeastyle} = [color=g31, mark=otimes*, line width=1pt, opacity=\getval{lineopacity}]
\tikzstyle{getefstyle} = [dashed, mark options=solid, color=g31, mark=otimes, line width=1pt, opacity=\getval{lineopacity}]

%% file: utils/plot_commands.tex
\newcommand{\corrplot}[1]{

	\pgfplotstableread{\getval{tefroot}/#1/\getval{mf_pcc}.tex}\tefpcctable
	
	\pgfplotstableread{\getval{mlmdistilfastroot}/#1/\getval{mf_pcc}.tex}\beapcctable

	\begin{tikzpicture}
		\begin{axis}[
			width=\plotwidth,
			height=\plotheight,
			xmin=\getval{x_min},
			xmax=\getval{x_max},
			ymin=-1.0,
			ymax=1.0,
			ymajorgrids=true,
			xmajorgrids=true,
			xtick={0.5*\getval{x_max}, \getval{x_max}},
			xticklabels={\getval{x_mid_tick}, $\displaystyle \rho$},
			grid style=dashed,
			legend pos=south west,
			legend style={nodes={scale=0.5, transform shape}},
		]

		\addplot[style=pccbeastyle] table[x=x, y=pcc_\getval{line_metric}]{\beapcctable};
		\addplot[name path=lq, style=pccbeastyle, opacity=0, forget plot] table[x={x}, y={pcc_\getval{fill_metric_low}}] {\beapcctable};
	    \addplot[name path=uq, style=pccbeastyle, opacity=0, forget plot] table[x={x}, y={pcc_\getval{fill_metric_high}}] {\beapcctable};
	    \addplot[style=pccbeastyle, fill opacity=\getval{fillopacity},, forget plot] fill between[of=lq and uq];
	    
		\addplot[style=pcctefstyle] table[x=x, y=pcc_\getval{line_metric}]{\tefpcctable};
		\addplot[name path=lq, style=pcctefstyle, opacity=0, forget plot] table[x={x}, y={pcc_\getval{fill_metric_low}}] {\tefpcctable};
	    \addplot[name path=uq, style=pcctefstyle, opacity=0, forget plot] table[x={x}, y={pcc_\getval{fill_metric_high}}] {\tefpcctable};
	    \addplot[style=pcctefstyle, fill opacity=\getval{fillopacity}, forget plot] fill between[of=lq and uq];
				\pgfplotstableclear{\tefpcctable}
		\pgfplotstableclear{\beapcctable}

		\end{axis}
	\end{tikzpicture}
}

\newcommand{\stsplot}[1]{

	\pgfplotstableread{\getval{tefroot}/#1/\getval{mf_sts_use}.tex}\tefusetable
	\pgfplotstableread{\getval{tefroot}/#1/\getval{mf_sts_tse}.tex}\teftsetable
	\pgfplotstableread{\getval{mlmdistilfastroot}/#1/\getval{mf_sts_use}.tex}\beausetable
	\pgfplotstableread{\getval{mlmdistilfastroot}/#1/\getval{mf_sts_tse}.tex}\beatsetable

	\begin{tikzpicture}
		\begin{axis}[
			width=\plotwidth,
			height=\plotheight,
			xmin=\getval{x_min},
			xmax=\getval{x_max},
			ymin=\getval{y_min},
			ymax=\getval{y_max},
			ymajorgrids=true,
			xmajorgrids=true,
			ytick={0, 0.6, 0.8, 1},
			yticklabels={0, 0.6, 0.8, 1},
			xtick={0.5*\getval{x_max}, \getval{x_max}},
			xticklabels={\getval{x_mid_tick}, $\displaystyle \rho$},
			grid style=dashed,
			legend pos=south west,
			legend style={nodes={scale=0.5, transform shape}},
		]
		
		\addplot[style=stsusebeastyle] table[x=x, y=sts_use_\getval{line_metric}]{\beausetable};
		\addplot[name path=lq, style=stsusebeastyle, opacity=0, forget plot] table[x={x}, y={sts_use_\getval{fill_metric_low}}] {\beausetable};
	    \addplot[name path=uq, style=stsusebeastyle, opacity=0, forget plot] table[x={x}, y={sts_use_\getval{fill_metric_high}}] {\beausetable};
	    \addplot[style=stsusebeastyle, fill opacity=\getval{fillopacity}, forget plot] fill between[of=lq and uq];

		\addplot[style=stsusetefstyle] table[x=x, y=sts_use_\getval{line_metric}]{\tefusetable};
		\addplot[name path=lq, style=stsusetefstyle, opacity=0, forget plot] table[x={x}, y={sts_use_\getval{fill_metric_low}}] {\tefusetable};
	    \addplot[name path=uq, style=stsusetefstyle, opacity=0, forget plot] table[x={x}, y={sts_use_\getval{fill_metric_high}}] {\tefusetable};
	    \addplot[style=stsusetefstyle, fill opacity=\getval{fillopacity}, forget plot] fill between[of=lq and uq];
		\addplot[style=ststsebeastyle] table[x=x, y=sts_tse_\getval{line_metric}]{\beatsetable};
		\addplot[name path=lq, style=ststsebeastyle, opacity=0, forget plot] table[x={x}, y={sts_tse_\getval{fill_metric_low}}] {\beatsetable};
	    \addplot[name path=uq, style=ststsebeastyle, opacity=0, forget plot] table[x={x}, y={sts_tse_\getval{fill_metric_high}}] {\beatsetable};
	    \addplot[style=ststsebeastyle, fill opacity=\getval{fillopacity}, forget plot] fill between[of=lq and uq];

		\addplot[style=ststsetefstyle] table[x=x, y=sts_tse_\getval{line_metric}]{\teftsetable};
		\addplot[name path=lq, style=ststsetefstyle, opacity=0, forget plot] table[x={x}, y={sts_tse_\getval{fill_metric_low}}] {\teftsetable};
	    \addplot[name path=uq, style=ststsetefstyle, opacity=0, forget plot] table[x={x}, y={sts_tse_\getval{fill_metric_high}}] {\teftsetable};
	    \addplot[style=ststsetefstyle, fill opacity=\getval{fillopacity}, forget plot] fill between[of=lq and uq];
		\pgfplotstableclear{\tefusetable}
		\pgfplotstableclear{\teftsetable}
		\pgfplotstableclear{\beausetable}
		\pgfplotstableclear{\beatsetable}

		\end{axis}
	\end{tikzpicture}
}

\newcommand{\ppplot}[1]{

	\pgfplotstableread{\getval{tefroot}/#1/\getval{mf_pp_r_d}.tex}\tefpptable
	\pgfplotstableread{\getval{mlmdistilfastroot}/#1/\getval{mf_pp_r_d}.tex}\beapptable

	\pgfplotstableread{\getval{tefroot}/#1/\getval{mf_orig_ge}.tex}\tefgetable
	\pgfplotstableread{\getval{mlmdistilfastroot}/#1/\getval{mf_orig_ge}.tex}\beagetable

	\begin{tikzpicture}
		\begin{axis}[
			width=\plotwidth,
			height=\plotheight,
			xmin=\getval{x_min},
			xmax=\getval{x_max},
			ymin=\getval{y_min},
			ymax=\getval{y_max},
			ymajorgrids=true,
			xmajorgrids=true,
			xtick={0.5*\getval{x_max}, \getval{x_max}},
			xticklabels={\getval{x_mid_tick}, $\displaystyle \rho$},
			grid style=dashed,
			legend pos=south east,
			legend style={nodes={scale=0.5, transform shape}},
		]

		\addplot[style=ppbeastyle] table[x=x, y=pp_r_d_\getval{line_metric}]{\beapptable};
		\addplot[name path=lq, style=ppbeastyle, opacity=0, forget plot] table[x={x}, y={pp_r_d_\getval{fill_metric_low}}] {\beapptable};
	    \addplot[name path=uq, style=ppbeastyle, opacity=0, forget plot] table[x={x}, y={pp_r_d_\getval{fill_metric_high}}] {\beapptable};
	    \addplot[style=ppbeastyle, fill opacity=\getval{fillopacity}, forget plot] fill between[of=lq and uq];
	    
		\addplot[style=pptefstyle] table[x=x, y=pp_r_d_\getval{line_metric}]{\tefpptable};
		\addplot[name path=lq, style=pptefstyle, opacity=0, forget plot] table[x={x}, y={pp_r_d_\getval{fill_metric_low}}] {\tefpptable};
	    \addplot[name path=uq, style=pptefstyle, opacity=0, forget plot] table[x={x}, y={pp_r_d_\getval{fill_metric_high}}] {\tefpptable};
	    \addplot[style=pptefstyle, fill opacity=\getval{fillopacity}, forget plot] fill between[of=lq and uq];
	    
	    \addplot[style=gebeastyle] table[x=x, y expr=\thisrow{adv_ge_mean}-\thisrow{orig_ge_mean}]{\beagetable};
		\addplot[name path=lq, style=gebeastyle, opacity=0, forget plot] table[x={x}, y expr=\thisrow{adv_ge_mean}-\thisrow{orig_ge_mean}-\thisrow{orig_ge_std}] {\beagetable};
	    \addplot[name path=uq, style=gebeastyle, opacity=0, forget plot] table[x={x}, y expr=\thisrow{adv_ge_mean}-\thisrow{orig_ge_mean}+\thisrow{orig_ge_std}] {\beagetable};
	    \addplot[style=gebeastyle, fill opacity=\getval{fillopacity}, forget plot] fill between[of=lq and uq];
	    
        \addplot[style=getefstyle] table[x=x, y expr=\thisrow{adv_ge_mean}-\thisrow{orig_ge_mean}]{\tefgetable};
		\addplot[name path=lq, style=getefstyle, opacity=0, forget plot] table[x={x}, y expr=\thisrow{adv_ge_mean}-\thisrow{orig_ge_mean}-\thisrow{orig_ge_std}] {\tefgetable};
	    \addplot[name path=uq, style=getefstyle, opacity=0, forget plot] table[x={x}, y expr=\thisrow{adv_ge_mean}-\thisrow{orig_ge_mean}+\thisrow{orig_ge_std}] {\tefgetable};
	    \addplot[style=getefstyle, fill opacity=\getval{fillopacity}, forget plot] fill between[of=lq and uq];

		\pgfplotstableclear{\tefpptable}
		\pgfplotstableclear{\beapptable}
		
		\pgfplotstableclear{\tefgetable}
		\pgfplotstableclear{\beagetable}

		\end{axis}
	\end{tikzpicture}
}

\newcommand{\kplot}[1]{

	\pgfplotstableread{\getval{tefroot}/#1/\getval{mf_et_use}.tex}\tefkusetable
	\pgfplotstableread{\getval{tefroot}/#1/\getval{mf_et_tse}.tex}\tefktsetable
	\pgfplotstableread{\getval{tefroot}/#1/\getval{mf_et_p}.tex}\tefkpptable
	
	\pgfplotstableread{\getval{mlmdistilfastroot}/#1/\getval{mf_et_use}.tex}\beakusetable
	\pgfplotstableread{\getval{mlmdistilfastroot}/#1/\getval{mf_et_tse}.tex}\beaktsetable
	\pgfplotstableread{\getval{mlmdistilfastroot}/#1/\getval{mf_et_p}.tex}\beakpptable
	
	\begin{tikzpicture}
		\begin{axis}[
			width=\plotwidth,
			height=\plotheight,
			xmin=\getval{x_min},
			xmax=\getval{x_max},
			ymin=\getval{y_min},
			ymax=\getval{y_max},
			ymajorgrids=true,
			xmajorgrids=true,
			xtick={0.5*\getval{x_max}, \getval{x_max}},
			xticklabels={\getval{x_mid_tick}, $\displaystyle \rho$},
			grid style=dashed,
			legend pos=north east,
			legend style={nodes={scale=0.5, transform shape}},
		]

		\addplot[style=kusebeastyle] table[x=x, y=et_use_\getval{line_metric}]{\beakusetable};
		\addplot[name path=lq, style=kusebeastyle, opacity=0, forget plot] table[x={x}, y={et_use_\getval{fill_metric_low}}] {\beakusetable};
	    \addplot[name path=uq, style=kusebeastyle, opacity=0, forget plot] table[x={x}, y={et_use_\getval{fill_metric_high}}] {\beakusetable};
	    \addplot[style=kusebeastyle, fill opacity=\getval{fillopacity}, forget plot] fill between[of=lq and uq];
	    
		\addplot[style=kusetefstyle] table[x=x, y=et_use_\getval{line_metric}]{\tefkusetable};
		\addplot[name path=lq, style=kusetefstyle, opacity=0, forget plot] table[x={x}, y={et_use_\getval{fill_metric_low}}] {\tefkusetable};
	    \addplot[name path=uq, style=kusetefstyle, opacity=0, forget plot] table[x={x}, y={et_use_\getval{fill_metric_high}}] {\tefkusetable};
	    \addplot[style=kusetefstyle, fill opacity=\getval{fillopacity}, forget plot] fill between[of=lq and uq];

		\addplot[style=ktsebeastyle] table[x=x, y=et_tse_\getval{line_metric}]{\beaktsetable};
		\addplot[name path=lq, style=ktsebeastyle, opacity=0, forget plot] table[x={x}, y={et_tse_\getval{fill_metric_low}}] {\beaktsetable};
	    \addplot[name path=uq, style=ktsebeastyle, opacity=0, forget plot] table[x={x}, y={et_tse_\getval{fill_metric_high}}] {\beaktsetable};
	    \addplot[style=ktsebeastyle, fill opacity=\getval{fillopacity}, forget plot] fill between[of=lq and uq];
	    
		\addplot[style=ktsetefstyle] table[x=x, y=et_tse_\getval{line_metric}]{\tefktsetable};
		\addplot[name path=lq, style=ktsetefstyle, opacity=0, forget plot] table[x={x}, y={et_tse_\getval{fill_metric_low}}] {\tefktsetable};
	    \addplot[name path=uq, style=ktsetefstyle, opacity=0, forget plot] table[x={x}, y={et_tse_\getval{fill_metric_high}}] {\tefktsetable};
	    \addplot[style=ktsebeastyle, fill opacity=\getval{fillopacity}, forget plot] fill between[of=lq and uq];

		\addplot[style=kppbeastyle] table[x=x, y=et_p_\getval{line_metric}]{\beakpptable};
		\addplot[name path=lq, style=kppbeastyle, opacity=0, forget plot] table[x={x}, y={et_p_\getval{fill_metric_low}}] {\beakpptable};
	    \addplot[name path=uq, style=kppbeastyle, opacity=0, forget plot] table[x={x}, y={et_p_\getval{fill_metric_high}}] {\beakpptable};
	    \addplot[style=kppbeastyle, fill opacity=\getval{fillopacity}, forget plot] fill between[of=lq and uq];
	    
		\addplot[style=kpptefstyle] table[x=x, y=et_p_\getval{line_metric}]{\tefkpptable};
		\addplot[name path=lq, style=kpptefstyle, opacity=0, forget plot] table[x={x}, y={et_p_\getval{fill_metric_low}}] {\tefkpptable};
	    \addplot[name path=uq, style=kpptefstyle, opacity=0, forget plot] table[x={x}, y={et_p_\getval{fill_metric_high}}] {\tefkpptable};
	    \addplot[style=kpptefstyle, fill opacity=\getval{fillopacity}, forget plot] fill between[of=lq and uq];

		\pgfplotstableclear{\tefkusetable}
		\pgfplotstableclear{\tefktsetable}
		\pgfplotstableclear{\tefkpptable}
		
		\pgfplotstableclear{\beakusetable}
		\pgfplotstableclear{\beaktsetable}
		\pgfplotstableclear{\beakpptable}

		\end{axis}
	\end{tikzpicture}
}

\newcommand{\allplotlegend}{
    \begin{tikzpicture}
        \tiny
        \begin{axis}[
            hide axis, xmin=0, xmax=1, ymin=0, ymax=1,
            legend columns=8
        ]
            \addlegendimage{pccbeastyle}
            \addlegendentry{PCC TEA};
            \addlegendimage{kusebeastyle}
            \addlegendentry{$k_{USE}$ TEA};
            \addlegendimage{ktsebeastyle}
            \addlegendentry{$k_{MiniLM}$ TEA};
            \addlegendimage{kppbeastyle}
            \addlegendentry{$k_{PP}$ TEA};
            \addlegendimage{stsusebeastyle}
            \addlegendentry{$STS_{USE}$ TEA};
            \addlegendimage{ststsebeastyle}
            \addlegendentry{$STS_{MiniLM}$ TEA};
            \addlegendimage{ppbeastyle}
            \addlegendentry{$\Delta_{PP}$ TEA};
            \addlegendimage{gebeastyle}
            \addlegendentry{$GE$ TEA};

            \addlegendimage{pcctefstyle}
            \addlegendentry{PCC TEF};
            \addlegendimage{kusetefstyle}
            \addlegendentry{$k_{USE}$ TEF};
            \addlegendimage{ktsetefstyle}
            \addlegendentry{$k_{MiniLM}$ TEF};
            \addlegendimage{kpptefstyle}
            \addlegendentry{$k_{PP}$ TEF};
            \addlegendimage{stsusetefstyle}
            \addlegendentry{$STS_{USE}$ TEF};
            \addlegendimage{ststsetefstyle}
            \addlegendentry{$STS_{MiniLM}$ TEF};
            \addlegendimage{pptefstyle}
            \addlegendentry{$\Delta_{PP}$ TEF};
            \addlegendimage{getefstyle}
            \addlegendentry{$GE$ TEF};
            
        \end{axis}
    \end{tikzpicture}
}

\newcommand{\relativescatterplot}[3]{
	\pgfplotstableread{\getval{dataroot}/auc_files/relative_#1_aucs.tex}\firstattacktable
	
	\begin{tikzpicture}
		\begin{axis}[
			width=\plotwidth,
			height=\plotheight,
			ymin=\getval{y_min},
			ymax=\getval{y_max},
			ymajorgrids=true,
			xmajorgrids=true,
			xtick={0, 1, 2, 3, 4},
			xticklabels={\scriptsize AG's News, \scriptsize MR, \scriptsize IMDB, \scriptsize Yelp, \scriptsize Fake News},
			grid style=dashed,
			legend pos=north east,
			legend style={nodes={scale=0.5, transform shape}},
		]
		\addplot[only marks, color=#3, opacity=0.5] table[x=x, y=#2] {\firstattacktable};
		\end{axis}
		\end{tikzpicture}

	\pgfplotstableclear{\firstattacktable}
}
\newcommand{\timinglineplot}[3]{

	\pgfplotstableread{\getval{timingroot}/#1_#2_#3_8_0.1.tex}\timingtable
	\pgfplotstableread{\getval{timingroot}/#1_#2_#3_8_0.25.tex}\timingtableslow

	\begin{tikzpicture}
		\begin{axis}[
			width=\plotwidth,
			height=\plotheight,
			ymin=\getval{y_min},
			ymax=\getval{y_max},
			ymajorgrids=true,
			xmajorgrids=true,
			xtick={0, 1, 2, 3},
			xticklabels={\scriptsize (0), \scriptsize (1), \scriptsize (2), \scriptsize (3)},
			xticklabel style={align=left},
			ytick={0, 0.5*\getval{y_max}, \getval{y_max}},
			yticklabels={0, \getval{y_mid_tick}, $\displaystyle s$},
			grid style=dashed,
			legend pos=north east,
			legend style={nodes={scale=0.7, transform shape}},
		]
		\addplot[name path=at, only marks, color=rg54, line width=1pt, opacity=\getval{lineopacity}, error bars/.cd, y dir=both, y explicit, error bar style={thick}] table[x=x, y=mean, y error=std]{\timingtable};
				
		\addplot[name path=at, only marks, color=rg50, mark=triangle*, line width=1pt, opacity=\getval{lineopacity},error bars/.cd, y dir=both, y explicit, error bar style={thick}] table[x=x, y=mean, y error=std]{\timingtableslow};
		\legend{$\rho_{max}=0.1$, $\rho_{max}=0.25$}
	
		\end{axis}
	\end{tikzpicture}
	\pgfplotstableclear{\timingtable}
}

%% file: src/abstract.tex
\begin{abstract}%
Explanations are crucial parts of deep neural network (DNN) classifiers. In high stakes applications, faithful and robust explanations are important to understand and gain trust in DNN classifiers. However, recent work has shown that state-of-the-art attribution methods in text classifiers are susceptible to imperceptible adversarial perturbations that alter explanations significantly while maintaining the correct prediction outcome. If undetected, this can critically mislead the users of DNNs. Thus, it is crucial to understand the influence of such adversarial perturbations on the networks' explanations and their perceptibility. In this work, we establish a novel definition of attribution robustness (AR) in text classification, based on Lipschitz continuity. Crucially, it reflects both attribution change induced by adversarial input alterations and perceptibility of such alterations. Moreover, we introduce a wide set of text similarity measures to effectively capture locality between two text samples and imperceptibility of adversarial perturbations in text. We then propose our novel \textsc{TransformerExplanationAttack} (TEA), a strong adversary that provides a tight estimation for attribution robustness in text classification. TEA uses state-of-the-art language models to extract word substitutions that result in fluent, contextual adversarial samples. Finally, with experiments on several text classification architectures, we show that TEA consistently outperforms current state-of-the-art AR estimators, yielding perturbations that alter explanations to a greater extent while being more fluent and less perceptible.%
\end{abstract}%

%% file: src/intro.tex
\section{Introduction}%
\label{sec:intro}%
Attribution methods aim to give insight into causal relationships between deep neural networks' (DNNs) inputs and their outcome prediction. They are fundamental to unravel the black-box nature of DNNs and are widely used both in the image and natural language domain. Commonly used attributions like Saliency \citep{saliencymap} Integrated Gradients \citep{integratedgradients}, DeepLift \citep{deeplift} and self-attention \citep{attentionmechanism} highlight input features that are deemed important for the DNNs in the inference process.\par%
\input{figures/example.tex}%
However, it has been shown recently that these methods do not deliver trustworthy, \textit{faithful} explanations \citep{interpretationfragile,faithfulness}. In particular, many of these attributions lack robustness towards small input perturbations. Carefully crafted, \textit{imperceptible} input alterations change the explanations significantly without modifying the output prediction of the DNNs. This violates the \textit{prediction assumption} of faithful explanations \citep{faithfulness}, i.e similar inputs having similar explanations for the same outputs. Figure \ref{fig:fragilityexample} exemplifies this fragility of attributions in text. In many safety-critical natural language processing problems, such as EHR classification \citep{aida}, adversarial robustness is a key factor for DNNs to be deployed in real life. For instance, a medical professional assessing EHRs would neither understand nor trust a model that yields two significantly different explanations for seemingly identical input texts and predictions. Hence, it is fundamental to understand how the networks and attributions behave in the presence of input perturbations and how perceptible those alterations are to the user.\par%
In this work, we focus on estimating the adversarial robustness of attribution maps (AR) in \textit{text classification} problems. Specifically, we are interested in investigating and quantifying the extent to which small input perturbations can alter explanations in DNNs and how perceptible such alterations are. Moreover, we focus on methods to find the optimal perturbations that maximize the change in attributions while being as imperceptible as possible. We summarize our contributions as follows:%
\begin{itemize}%
  \item We are the first to introduce a definition of attribution robustness (AR) in text classification, derived from the Lipschitz-constant, that takes into account both the attribution distance and perceptibility of perturbations.%
  \item We are the first to propose a diverse set of metrics to capture numerous aspects of perceptibility of small input perturbations in text.%
  \item We introduce a novel and powerful attack technique, \textsc{TransformerExplanationAttack} (TEA), that we show consistently outperforms state-of-the-art adversaries and therefore allows us to more tightly estimate attribution robustness in text classifiers.%
  \item We are the first to utilize masked language models (MLMS) for context-aware candidate extraction in attribution robustness estimation. This is crucial, as domain-specific MLMs are becoming increasingly available, making them a progressively attractive alternative to custom synonym embeddings on which current methods have to rely.
  \item We successfully speed up robustness estimation with the usage of distilled language models and batch masking.%
\end{itemize}%

%% file: figures/example.tex
\begin{figure*}[t]%
\begin{tabular}{ p{0.3\textwidth} | p{0.3\textwidth} | p{0.3\textwidth}}%
\hline%
%
%
\centering \textbf{Original sample}%
&%
\centering \textbf{TEA perturbed sample}{\newline}(ours)%
&%
\centering\arraybackslash \textbf{TEF perturbed sample}{\newline}\citep{ivankay2021fooling}%
\\%
\hline%
\hline%
%
%
\small%
\textcolor[rgb]{0.26,0.58,0.76}{\textbf{peek }}\textcolor[rgb]{0,0,0}{at }\textcolor[rgb]{0,0,0}{the }\textcolor[rgb]{0,0,0}{week }\textcolor[rgb]{0,0,0}{: }\textcolor[rgb]{0.84,0.38,0.30}{\textbf{ben }}\textcolor[rgb]{0,0,0}{vs. }\textcolor[rgb]{0,0,0}{the }\textcolor[rgb]{0.84,0.38,0.30}{\textbf{streak }}\textcolor[rgb]{0.26,0.58,0.76}{\textbf{| }}\textcolor[rgb]{0,0,0}{yet }\textcolor[rgb]{0,0,0}{another }\textcolor[rgb]{0,0,0}{risky }\textcolor[rgb]{0.84,0.38,0.30}{\textbf{game }}\textcolor[rgb]{0,0,0}{for }\textcolor[rgb]{0,0,0}{that }\textcolor[rgb]{0,0,0}{patriots }\textcolor[rgb]{0,0,0}{winning }\textcolor[rgb]{0,0,0}{streak }\textcolor[rgb]{0,0,0}{, }\textcolor[rgb]{0,0,0}{now }\textcolor[rgb]{0,0,0}{at }\textcolor[rgb]{0,0,0}{21 }\textcolor[rgb]{0,0,0}{. }\textcolor[rgb]{0,0,0}{pittsburgh }\textcolor[rgb]{0,0,0}{hasn }\textcolor[rgb]{0,0,0}{\# }\textcolor[rgb]{0,0,0}{39;t }\textcolor[rgb]{0,0,0}{lost }\textcolor[rgb]{0,0,0}{at }\textcolor[rgb]{0,0,0}{home }\textcolor[rgb]{0,0,0}{, }\textcolor[rgb]{0,0,0}{and }\textcolor[rgb]{0.7,0.09,0.17}{\textbf{rookie }}\textcolor[rgb]{0.84,0.38,0.30}{\textbf{quarterback }}\textcolor[rgb]{0,0,0}{ben }\textcolor[rgb]{0,0,0}{roethlisberger }\textcolor[rgb]{0,0,0}{hasn }\textcolor[rgb]{0,0,0}{\# }\textcolor[rgb]{0,0,0}{39;t }\textcolor[rgb]{0,0,0}{lost }\textcolor[rgb]{0,0,0}{, }\textcolor[rgb]{0,0,0}{period }\textcolor[rgb]{0,0,0}{. }%
&%
\small%
\textcolor[rgb]{0.26,0.58,0.76}{\textbf{peek }}\textcolor[rgb]{0,0,0}{at }\textcolor[rgb]{0,0,0}{the }\textcolor[rgb]{0,0,0}{{\underline{playoffs}} }\textcolor[rgb]{0,0,0}{: }\textcolor[rgb]{0,0,0}{ben }\textcolor[rgb]{0,0,0}{vs. }\textcolor[rgb]{0,0,0}{the }\textcolor[rgb]{0.13,0.40,0.67}{\textbf{{\underline{steelers}} }}\textcolor[rgb]{0,0,0}{| }\textcolor[rgb]{0.84,0.38,0.30}{\textbf{yet }}\textcolor[rgb]{0,0,0}{another }\textcolor[rgb]{0,0,0}{risky }\textcolor[rgb]{0.13,0.40,0.67}{\textbf{game }}\textcolor[rgb]{0,0,0}{for }\textcolor[rgb]{0,0,0}{that }\textcolor[rgb]{0,0,0}{patriots }\textcolor[rgb]{0,0,0}{winning }\textcolor[rgb]{0.26,0.58,0.76}{\textbf{streak }}\textcolor[rgb]{0,0,0}{, }\textcolor[rgb]{0,0,0}{now }\textcolor[rgb]{0,0,0}{at }\textcolor[rgb]{0,0,0}{21 }\textcolor[rgb]{0.26,0.58,0.76}{\textbf{. }}\textcolor[rgb]{0.26,0.58,0.76}{\textbf{pittsburgh }}\textcolor[rgb]{0,0,0}{hasn }\textcolor[rgb]{0,0,0}{\# }\textcolor[rgb]{0.26,0.58,0.76}{\textbf{{\underline{34}} }}\textcolor[rgb]{0,0,0}{lost }\textcolor[rgb]{0,0,0}{at }\textcolor[rgb]{0,0,0}{home }\textcolor[rgb]{0,0,0}{, }\textcolor[rgb]{0,0,0}{and }\textcolor[rgb]{0,0,0}{rookie }\textcolor[rgb]{0.84,0.38,0.30}{\textbf{quarterback }}\textcolor[rgb]{0.26,0.58,0.76}{\textbf{ben }}\textcolor[rgb]{0,0,0}{roethlisberger }\textcolor[rgb]{0,0,0}{hasn }\textcolor[rgb]{0,0,0}{\# }\textcolor[rgb]{0,0,0}{39;t }\textcolor[rgb]{0,0,0}{lost }\textcolor[rgb]{0,0,0}{, }\textcolor[rgb]{0,0,0}{period }\textcolor[rgb]{0,0,0}{{\underline{\ensuremath{>}}} }%
&%
\small%
\textcolor[rgb]{0,0,0}{{\underline{hoodwink}} }\textcolor[rgb]{0,0,0}{at }\textcolor[rgb]{0.84,0.38,0.30}{\textbf{the }}\textcolor[rgb]{0.7,0.09,0.17}{\textbf{{\underline{zou}} }}\textcolor[rgb]{0,0,0}{: }\textcolor[rgb]{0.26,0.58,0.76}{\textbf{{\underline{suis}} }}\textcolor[rgb]{0.7,0.09,0.17}{\textbf{vs. }}\textcolor[rgb]{0,0,0}{the }\textcolor[rgb]{0.26,0.58,0.76}{\textbf{{\underline{wave}} }}\textcolor[rgb]{0.13,0.40,0.67}{\textbf{| }}\textcolor[rgb]{0.84,0.38,0.30}{\textbf{yet }}\textcolor[rgb]{0,0,0}{another }\textcolor[rgb]{0.26,0.58,0.76}{\textbf{risky }}\textcolor[rgb]{0,0,0}{game }\textcolor[rgb]{0.26,0.58,0.76}{\textbf{for }}\textcolor[rgb]{0,0,0}{that }\textcolor[rgb]{0.84,0.38,0.30}{\textbf{patriots }}\textcolor[rgb]{0,0,0}{winning }\textcolor[rgb]{0.26,0.58,0.76}{\textbf{streak }}\textcolor[rgb]{0,0,0}{, }\textcolor[rgb]{0,0,0}{now }\textcolor[rgb]{0.26,0.58,0.76}{\textbf{at }}\textcolor[rgb]{0,0,0}{21 }\textcolor[rgb]{0.13,0.40,0.67}{\textbf{. }}\textcolor[rgb]{0.13,0.40,0.67}{\textbf{pittsburgh }}\textcolor[rgb]{0.84,0.38,0.30}{\textbf{hasn }}\textcolor[rgb]{0,0,0}{\# }\textcolor[rgb]{0,0,0}{39;t }\textcolor[rgb]{0,0,0}{lost }\textcolor[rgb]{0,0,0}{at }\textcolor[rgb]{0.26,0.58,0.76}{\textbf{home }}\textcolor[rgb]{0,0,0}{, }\textcolor[rgb]{0.26,0.58,0.76}{\textbf{and }}\textcolor[rgb]{0.84,0.38,0.30}{\textbf{rookie }}\textcolor[rgb]{0.84,0.38,0.30}{\textbf{quarterback }}\textcolor[rgb]{0.26,0.58,0.76}{\textbf{ben }}\textcolor[rgb]{0,0,0}{roethlisberger }\textcolor[rgb]{0,0,0}{hasn }\textcolor[rgb]{0,0,0}{\# }\textcolor[rgb]{0,0,0}{39;t }\textcolor[rgb]{0,0,0}{lost }\textcolor[rgb]{0,0,0}{, }\textcolor[rgb]{0,0,0}{period }\textcolor[rgb]{0,0,0}{. }%
\\[-0.2cm]%
\centering \small F$(\vs, {\mathrm{\mathrm{``Sports"}}) {}={} 0.99}$%
&%
\centering \small F$(\vs, {\mathrm{\mathrm{``Sports"}}) {}={} 0.95}$%
&%
\centering\arraybackslash \small F$(\vs, {\mathrm{\mathrm{``Sports"}}) {}={} 1.0}$%
\\[0.0cm]%
&%
\centering \small \textbf{PCC}: 0.02%
&%
\centering\arraybackslash \small \textbf{PCC}: 0.22%
\\[0.0cm]%
&%
\centering \small \textit{SemS}: 0.97, \textit{k}: 14.9%
&%
\centering\arraybackslash \small \textit{SemS}: 0.9, \textit{k}: 3.4%
\\%
\hline%
%
%
%
\small%
\textcolor[rgb]{0.7,0.09,0.17}{\textbf{press }}\textcolor[rgb]{0.84,0.38,0.30}{\textbf{the }}\textcolor[rgb]{0.84,0.38,0.30}{\textbf{delete }}\textcolor[rgb]{0.13,0.40,0.67}{\textbf{key }}\textcolor[rgb]{0,0,0}{. }%
&%
\small%
\textcolor[rgb]{0.13,0.40,0.67}{\textbf{{\underline{hit}} }}\textcolor[rgb]{0,0,0}{the }\textcolor[rgb]{0,0,0}{delete }\textcolor[rgb]{0.26,0.58,0.76}{\textbf{key }}\textcolor[rgb]{0.84,0.38,0.30}{\textbf{. }}%
&%
\small%
\textcolor[rgb]{0.26,0.58,0.76}{\textbf{{\underline{newspaper}} }}\textcolor[rgb]{0.7,0.09,0.17}{\textbf{the }}\textcolor[rgb]{0.7,0.09,0.17}{\textbf{delete }}\textcolor[rgb]{0.13,0.40,0.67}{\textbf{key }}\textcolor[rgb]{0.84,0.38,0.30}{\textbf{. }}%
\\[0.15cm]%
\centering \small F$(\vs, {\mathrm{\mathrm{``Negative"}}) {}={} 0.99}$%
&%
\centering \small F$(\vs, {\mathrm{\mathrm{``Negative"}}) {}={} 0.95}$
&%
\centering\arraybackslash \small F$(\vs, {\mathrm{\mathrm{``Negative"}}) {}={} 0.95}$
\\[0.0cm]%
&%
\centering \small \textbf{PCC}: -0.05%
&%
\centering\arraybackslash \small \textbf{PCC}: 0.6%
\\[0.0cm]%
&%
\centering \small \textit{SemS}: 0.98, \textit{k}: 30%
&%
\centering\arraybackslash \small \textit{SemS}: 0.8, \textit{k}: 1.1%
\\%
\hline%
%
%
\small%
\textcolor[rgb]{0,0,0}{intel }\textcolor[rgb]{0,0,0}{seen }\textcolor[rgb]{0.84,0.38,0.30}{\textbf{readying }}\textcolor[rgb]{0,0,0}{new }\textcolor[rgb]{0.84,0.38,0.30}{\textbf{wi }}\textcolor[rgb]{0,0,0}{- }\textcolor[rgb]{0.7,0.09,0.17}{\textbf{fi }}\textcolor[rgb]{0,0,0}{chips }\textcolor[rgb]{0,0,0}{| }\textcolor[rgb]{0,0,0}{intel }\textcolor[rgb]{0,0,0}{corp }\textcolor[rgb]{0,0,0}{. }\textcolor[rgb]{0,0,0}{this }\textcolor[rgb]{0,0,0}{week }\textcolor[rgb]{0,0,0}{isexpected }\textcolor[rgb]{0,0,0}{to }\textcolor[rgb]{0,0,0}{introduce }\textcolor[rgb]{0,0,0}{a }\textcolor[rgb]{0,0,0}{chip }\textcolor[rgb]{0,0,0}{that }\textcolor[rgb]{0,0,0}{adds }\textcolor[rgb]{0,0,0}{support }\textcolor[rgb]{0,0,0}{for }\textcolor[rgb]{0,0,0}{a }\textcolor[rgb]{0,0,0}{relativelyobscure }\textcolor[rgb]{0,0,0}{version }\textcolor[rgb]{0,0,0}{of }\textcolor[rgb]{0,0,0}{wi }\textcolor[rgb]{0,0,0}{- }\textcolor[rgb]{0.84,0.38,0.30}{\textbf{fi }}\textcolor[rgb]{0,0,0}{, }\textcolor[rgb]{0,0,0}{analysts }\textcolor[rgb]{0,0,0}{said }\textcolor[rgb]{0,0,0}{on }\textcolor[rgb]{0,0,0}{monday }\textcolor[rgb]{0,0,0}{, }\textcolor[rgb]{0,0,0}{in }\textcolor[rgb]{0,0,0}{a }\textcolor[rgb]{0,0,0}{movethat }\textcolor[rgb]{0,0,0}{could }\textcolor[rgb]{0,0,0}{help }\textcolor[rgb]{0,0,0}{ease }\textcolor[rgb]{0,0,0}{congestion }\textcolor[rgb]{0,0,0}{on }\textcolor[rgb]{0.84,0.38,0.30}{\textbf{wireless }}\textcolor[rgb]{0,0,0}{networks }\textcolor[rgb]{0,0,0}{. }%
&%
\small%
\textcolor[rgb]{0,0,0}{intel }\textcolor[rgb]{0,0,0}{seen }\textcolor[rgb]{0.84,0.38,0.30}{\textbf{readying }}\textcolor[rgb]{0.7,0.09,0.17}{\textbf{{\underline{wireless}} }}\textcolor[rgb]{0,0,0}{wi }\textcolor[rgb]{0,0,0}{- }\textcolor[rgb]{0.84,0.38,0.30}{\textbf{fi }}\textcolor[rgb]{0,0,0}{chips }\textcolor[rgb]{0,0,0}{| }\textcolor[rgb]{0,0,0}{intel }\textcolor[rgb]{0,0,0}{corp }\textcolor[rgb]{0,0,0}{. }\textcolor[rgb]{0,0,0}{this }\textcolor[rgb]{0,0,0}{week }\textcolor[rgb]{0,0,0}{isexpected }\textcolor[rgb]{0,0,0}{to }\textcolor[rgb]{0,0,0}{{\underline{launch}} }\textcolor[rgb]{0,0,0}{a }\textcolor[rgb]{0,0,0}{{\underline{specification}} }\textcolor[rgb]{0,0,0}{that }\textcolor[rgb]{0,0,0}{{\underline{added}} }\textcolor[rgb]{0,0,0}{support }\textcolor[rgb]{0,0,0}{for }\textcolor[rgb]{0,0,0}{a }\textcolor[rgb]{0,0,0}{relativelyobscure }\textcolor[rgb]{0,0,0}{version }\textcolor[rgb]{0,0,0}{of }\textcolor[rgb]{0,0,0}{wi }\textcolor[rgb]{0,0,0}{- }\textcolor[rgb]{0,0,0}{fi }\textcolor[rgb]{0,0,0}{, }\textcolor[rgb]{0,0,0}{analysts }\textcolor[rgb]{0,0,0}{said }\textcolor[rgb]{0,0,0}{on }\textcolor[rgb]{0,0,0}{monday }\textcolor[rgb]{0,0,0}{, }\textcolor[rgb]{0,0,0}{in }\textcolor[rgb]{0,0,0}{a }\textcolor[rgb]{0,0,0}{movethat }\textcolor[rgb]{0,0,0}{could }\textcolor[rgb]{0,0,0}{help }\textcolor[rgb]{0,0,0}{ease }\textcolor[rgb]{0,0,0}{congestion }\textcolor[rgb]{0,0,0}{on }\textcolor[rgb]{0,0,0}{wireless }\textcolor[rgb]{0,0,0}{networks }\textcolor[rgb]{0,0,0}{. }%
&
\small%
\textcolor[rgb]{0.84,0.38,0.30}{\textbf{intel }}\textcolor[rgb]{0,0,0}{seen }\textcolor[rgb]{0,0,0}{readying }\textcolor[rgb]{0,0,0}{{\underline{nouveau}} }\textcolor[rgb]{0.7,0.09,0.17}{\textbf{wi }}\textcolor[rgb]{0,0,0}{- }\textcolor[rgb]{0.7,0.09,0.17}{\textbf{fi }}\textcolor[rgb]{0,0,0}{chips }\textcolor[rgb]{0,0,0}{| }\textcolor[rgb]{0,0,0}{intel }\textcolor[rgb]{0,0,0}{corp }\textcolor[rgb]{0,0,0}{. }\textcolor[rgb]{0,0,0}{this }\textcolor[rgb]{0,0,0}{week }\textcolor[rgb]{0,0,0}{isexpected }\textcolor[rgb]{0,0,0}{to }\textcolor[rgb]{0.26,0.58,0.76}{\textbf{{\underline{insert}} }}\textcolor[rgb]{0,0,0}{a }\textcolor[rgb]{0,0,0}{{\underline{dies}} }\textcolor[rgb]{0,0,0}{that }\textcolor[rgb]{0,0,0}{{\underline{summing}} }\textcolor[rgb]{0,0,0}{support }\textcolor[rgb]{0,0,0}{for }\textcolor[rgb]{0,0,0}{a }\textcolor[rgb]{0,0,0}{relativelyobscure }\textcolor[rgb]{0,0,0}{version }\textcolor[rgb]{0,0,0}{of }\textcolor[rgb]{0,0,0}{wi }\textcolor[rgb]{0,0,0}{- }\textcolor[rgb]{0.7,0.09,0.17}{\textbf{fi }}\textcolor[rgb]{0,0,0}{, }\textcolor[rgb]{0.26,0.58,0.76}{\textbf{analysts }}\textcolor[rgb]{0,0,0}{said }\textcolor[rgb]{0,0,0}{on }\textcolor[rgb]{0,0,0}{monday }\textcolor[rgb]{0,0,0}{, }\textcolor[rgb]{0,0,0}{in }\textcolor[rgb]{0,0,0}{a }\textcolor[rgb]{0,0,0}{movethat }\textcolor[rgb]{0,0,0}{could }\textcolor[rgb]{0,0,0}{help }\textcolor[rgb]{0,0,0}{ease }\textcolor[rgb]{0,0,0}{congestion }\textcolor[rgb]{0,0,0}{on }\textcolor[rgb]{0.7,0.09,0.17}{\textbf{wireless }}\textcolor[rgb]{0,0,0}{networks }\textcolor[rgb]{0,0,0}{. }%
\\[0.3cm]%
\centering \small F($\vs, {\mathrm{\mathrm{``Sci / Tech"}}) {}={} 0.78}$%
&%
\centering \small F$(\vs, {\mathrm{\mathrm{``Sci / Tech"}}) {}={} 0.95}$%
&%
\centering\arraybackslash \small F$(\vs, {\mathrm{\mathrm{``Sci / Tech"}}) {}={} 0.95}$%
\\[0.0cm]%
&%
\centering \small \textbf{PCC}: 0.27%
&%
\centering\arraybackslash \small \textbf{PCC}: 0.28%
\\[0.0cm]%
&%
\small \centering \textit{SemS}: 0.98, \textit{k}: 20%
&%
\centering\arraybackslash \small \textit{SemS}: 0.91, \textit{k}: 4%
\\%
\hline%
%
%
\hline%
\end{tabular}%
\caption{Attribution maps in text sequence classifiers are not robust to word substitutions. By carefully altering the original sample, the perturbed samples result in significantly different attribution maps while maintaining the prediction confidence F in the correct class. Red words have positive attribution values, i.e. contribute \textit{towards} the true class, while blue words with negative attributions \textit{against} it. Our novel TEA attack yields perturbed samples that have lower \textit{Pearson Correlation Coefficient} (PCC) values between the attribution maps of original and perturbed inputs, as well as higher semantic similarity values (SemS) of the original and adversarial sentences, compared to the baseline TEF attack. This results in higher estimated robustness constants $k$ (see Section \ref{sec:methods}), thus lower robustness of the classifiers against the attack.}%
\label{fig:fragilityexample}%
\end{figure*}%

%% file: src/related_work.tex
\section{Related work}%
\label{sec:relwork}%
The robustness aspect of faithful explanations \citep{faithfulness} has recently been studied with increasing interest. The authors \citet{interpretationfragile} were the first to show that attribution methods like Integrated Gradients \citep{integratedgradients} and DeepLift \citep{deeplift}, amongst others, lack robustness to local, imperceptible perturbations in the input that lead to significantly altered attribution maps while maintaining the correct prediction of the image classifier. The works of \citet{geometryblame}, \citet{igsum}, \citet{curvatureregularization}, \citet{rigotti2022} and \citet{far} have further studied this phenomenon and established theoretical frameworks to understand and mitigate the lack of attribution robustness in the image domain.\par%
However, explanation robustness in natural language processing has not been explored as deeply. The authors \citet{attentionnotexplanation} and \citet{attentionnotnotexplanation} show that similar inputs can lead to similar attention values but different predictions, and that models can be retrained to yield different attention values for similar inputs and outputs. This however does not directly contradict the prediction assumption of faithfulness \citep{faithfulness} as discussed by \citet{attentionnotnotexplanation}. Closer to our work, the works \citet{ivankay2021fooling} and \citet{devilswork} are the first to prove that explanations in text classifiers are also susceptible to input changes in a very small local neighbourhood of the input. They introduce \textsc{TextExplanationFooler} (TEF) as a baseline to alter attributions and estimate local robustness of widely used attributions in text classification. However, the authors' definition of AR does not take semantic distances between original and adversarial samples into account. Thus their attack results in out-of-context and non-fluent adversarial samples, rendering such perturbations easier to detect, as it draws token substitution candidates from a separately trained custom synonym embedding.%
%

%% file: src/prelim.tex
\section{Preliminaries}%
\label{sec:prelims}%
A text dataset $\displaystyle \sS$ is comprised of $\displaystyle N$ text samples $\displaystyle \vs_i$, each containing a series of tokens $w_j$ from a vocabulary $\sW$ and true labels $l_i$ drawn from the label set $\sL$. A text classifier $F$ is a function that maps each sample $\vs_i$ to a label $y_i \in \sL$. It consists of an embedding function $E$ and a classifier function $f$. The embedding function $\displaystyle E: \sS \rightarrow \R^{d \times p} ,\; \displaystyle E(\vs) = \mX$ maps the text samples $\vs_i$ to a continuous embedding $\mX$, while the classifier function $\displaystyle f: \R^{d \times p} \rightarrow \R^{|\sL|},\; \displaystyle f(\displaystyle \mX) = \displaystyle \vo$ maps the embeddings to the output probabilities for each class.\par%
An \textit{attribution function} $\displaystyle A(\displaystyle \vs, F, \evl) = \displaystyle \va$ assigns a real number to each token $w_j$ in sample $\vs$. This represents the tokens influence towards the classification outcome. A positive value represents a token that is deemed relevant \textit{towards} the predicted label $y$, a negative value \textit{against} it. We consider the attributions methods Saliency (S) \citep{saliencymap}, Integrated Gradients (IG) \citep{integratedgradients} and Attention (A) \citep{attentionmechanism}.\par%
The \textit{perplexity} \citep{perplexity} of a text sample $\vs$ with tokens $w_j$ given a language model $L$ measures how well the probability distribution given by $L$ predicts sample $\vs$, as defined in Equation (\ref{eqn:perplexity}):%
\begin{equation}%
\label{eqn:perplexity}
    PP(\vs | L) = 2^{-\sum_{w_j \in \vs}p(w_j|L, \vs)\log p(w_j|L, \vs)}%
\end{equation}%
where $PP$ denotes the perplexity of the text sample $\vs$ and $p(w_j|L, \vs)$ the probability of token $w_j$ given $L$ and $\vs$. Low perplexity values indicate that the model $L$ has well captured the true distribution of the text dataset $\sS$.\par%
\textit{Sentence encoders} are embedding functions $E_s : \sS \rightarrow \R^m ,\; \displaystyle E_s(\vs) = \vm$ that assign a continuous embedding vector of dimension $m$ to each text sample \citep{sbert}. These embeddings are used to capture higher-level representations of the text samples or short paragraphs that can be used to train downstream tasks effectively.%

%% file: src/methods.tex
\section{Attribution Robustness}%
\label{sec:methods}%
In this section, we introduce our new definition of attribution robustness (AR) in text classifiers. We describe our attribution and text distance measures. Furthermore, we introduce the optimization problem of estimating AR, our threat model as well as our novel estimator algorithm.%
\subsection{Attribution Robustness in Text}%
Most related work defines AR as the maximal attribution distance with a given locality constraint in the search space \citep{ivankay2021fooling, devilswork}. We argue that this is potentially problematic, as the perturbation size of the input is not taken into account. Two adversarial samples with similarly altered attributions might in fact strongly differ in terms of how well they maintain semantic similarity to the original sample (see e.g. $3^{\mathrm{rd}}$ example in Figure \ref{fig:fragilityexample}). This suggests that a proper measure of attribution robustness should ascribe higher robustness to methods that are only vulnerable to larger perturbation while being impervious to imperceptible ones. Thus, we define attribution robustness for a given text sample $\vs$ with true label $l$ as functions of both resulting attribution distance and input perturbation size, written in Equation (\ref{eqn:ar}).%
\begin{equation}%
\label{eqn:ar}%
    k(\vs) {}={} \max_{\tilde{\vs} \in N(\vs)} \frac{d \big[ A(\tilde{\vs}, F, l), \; A(\vs, F, l) \big]}{d_s(\tilde{\vs},\vs)}%
\end{equation}%
with the constraint that the predicted classes of $\tilde{\vs}$ and $\vs$ are equal, written in Equation (\ref{eqn:pred_constraint}).%
\begin{equation}%
\label{eqn:pred_constraint}%
    \argmax_{i \in \{1...|\sL|\}}F_i(\tilde{\vs}) {}={} \argmax_{i \in \{1...|\sL|\}}F_i(\vs)%
\end{equation}%
Here, $d$ denotes the distance between attribution maps $\displaystyle A(\tilde{\vs}, F, l)$ and $\displaystyle A(\vs, F, l)$, $F$ the sequence classifier with output probability $F_i$ for class $i$, and $\displaystyle d_s$ the distance of input text samples $\displaystyle \tilde{\vs}$ and $\displaystyle \vs$. $N(\vs)$ indicates a neighborhood of $\vs$: $N(\vs)=\tilde{\vs} \; | \; d_s(\tilde{\vs},\vs) < \varepsilon$ for a small $\varepsilon$. This definition is inspired by the Lipschitz continuity constant \citep{lipschitz} and the robustness assumption of faithful explanations \citep{faithfulness}. The estimated attribution robustness of an attribution method $\displaystyle A$ on a model $F$ then becomes the expected per-sample $k(\vs)$ on dataset $\sS$, see Equation (\ref{eqn:expectedar}).%
\begin{equation}%
\label{eqn:expectedar}%
    k(A, F) {}={} \mathbb{E}_{\vs \in \sS} \big[ k(\vs) \big]%
\end{equation}%
We call this $k$ the estimated \textit{Lipschitz attribution robustness (AR) constant}. The robustness of attribution method $A$ on the model $F$ is inversely proportional to $k(A, F)$, as high values correlate with large attribution distances and small input perturbations, which indicates low robustness.\par%
\subsection{Distances in Text Data}%
\label{subsec:distances}%
In order to compute the attribution robustness constants $k$ from Equation (\ref{eqn:expectedar}), the distance measures in the numerator and denominator of Equation (\ref{eqn:ar}) need to be defined. In explainable AI, it is often argued that only the relative rank between input features or tokens is important when explaining the outcome of a classifier, or even only the top-k features. Users frequently focus on the features deemed most important to explain a decision and disregard the less important ones \citep{far,interpretationfragile,geometryblame}. Therefore, it is common practice \citep{ivankay2021fooling,devilswork} to use correlation coefficients and top-k intersections as distance measures between attributions, as these tend to capture the human understanding of distance between explanations well. For this reason, we utilize the Pearson correlation coefficient (PCC) \citep{pcc} as attribution distance $\displaystyle d \big[ A(\tilde{\vs}, F, l), \; A(\vs, F, l) \big] {}={} 1 - \frac{1 + PCC\big[ A(\tilde{\vs}, F, l), \; A(\vs, F, l) \big]}{2}$ of Equation (\ref{eqn:ar}).\par%
Measuring distance between text inputs in the adversarial setting is not as straightforward as in the image domain, where $L_p$-norm induced distances are common. String distance metrics can only be used limitedly, as two words can have similar characters but entirely different meanings. For this reason, we propose the following set of measures to effectively capture the perturbation size in the denominator of Equation (\ref{eqn:ar}).\par%
First, we utilize pretrained sentence encoders to measure the semantic textual similarity between the original and adversarial text samples. This can be computed by the cosine similarity between the sentence embeddings of the two text samples, see Equation (\ref{eqn:semanticsim}).%
\begin{equation}%
\label{eqn:semanticsim}%
d_{\mathrm{s}}(\tilde{\vs},\vs) {}={} 1 - \frac{s_{cos}[E_s(\tilde{\vs}), E_s(\vs)] + 1}{2}%
\end{equation}%
where $d_{\mathrm{s}}$ denotes the semantic distance between samples $\tilde{\vs}$ and $\vs$, $s_{cos}$ the cosine similarity, and $E_s(\tilde{\vs})$ and $E_s(\vs)$ the sentence embeddings of the two input samples. The semantic textual similarity provides a measure how close the two inputs are in their semantic meaning. To this end, the Universal Sentence Encoder \citep{use} is widely-used in adversarial text setups \cite{advbert,ivankay2021fooling}. However, this architecture is not state-of-the-art on the STSBenchmark dataset \citep{stsbenchmark}, a benchmark used to evaluate semantic textual similarity. Therefore, we utilize a second sentence encoder architecture trained by the authors \citet{tse}, MiniLM. This model achieves close to state-of-the-art performance on the STSBenchmark while maintaining a low computational cost.\par%
Our second input distance is derived from the perplexity of original and adversarial inputs $\tilde{\vs}$ and $\vs$. We capture the relative increase of perplexity when perturbing the original sentence $\vs$, given the pretrained GPT-2 language model \citep{gpt2} (Equation \ref{eqn:ppincrease}).%
\begin{equation}
\label{eqn:ppincrease}
    d_{s}(\tilde{\vs},\vs) {}={} \frac{PP(\tilde{\vs}|L) - PP(\vs|L)}{PP(\vs|L) + \varepsilon}
\end{equation}
where $d_{\mathrm{s}}$ denotes the distance between inputs $\tilde{\vs}$ and $\vs$, $PP$ the perplexity of the text sample given the GPT-2 language model $L$, $\varepsilon$ a small constant.\par%
Lastly, we capture the increase of grammatical errors in the input samples using the LanguageTool API \footnote{https://languagetool.org}. As grammatical errors are easily perceived by the human observer, they significantly contribute to the perceptibility of adversarial perturbations \citep{hotflip}.%
\subsection{Robustness Estimation}%
Given our AR definition in Equation (\ref{eqn:ar}), in order to estimate the true robustness of an attribution method for a given model, all possible input sequences $\displaystyle \tilde{\vs}$ within the neighbourhood $N$ of $\vs$ would have to be checked. This is a computationally intractable problem. Therefore we restrict the search space (i.e. the neighbourhood $N$) to sequences $\tilde{\vs}$ that only contain token substitutions from the predefined vocabulary set $\sW$. Moreover, we restrict the ratio of substituted tokens in the original sequence to $\rho_{max}$, considering only $|\sC|$ number of possible substitutions for each token in $\vs$. The number $|\sC|$ is chosen to yield high attribution distance while keeping the computation cost low, detailed in Section \ref{sec:experiments}. This way, we reduce the total perturbation set from ${|\sW|}^{|\vs|}$ to ${|\sC|}^{|\vs| \cdot \rho_{max}}$ samples. The adversarial sequence $\vs_\mathrm{adv}$ then becomes the perturbed sequence that maximizes $k(\vs)$ from Equation (\ref{eqn:ar})\par%
\input{figures/algorithms.tex}%
We estimate AR with our novel \textsc{TransformerExplanationAttack} (TEA). TEA is a black-box attack, only having access to the model's prediction and the accompanying attributions, not the intermediate representations, architectural information or gradients. TEA consists of the following two steps.\par%
\paragraph{Step 1: Word importance ranking.} The first step extracts a priority ranking of tokens in the input text sample $\vs$. For each word $w_i$ in $\vs$, TEA computes $\displaystyle I_{\evw_i} {}={} d \big[ A(\vs_{w_i\rightarrow 0}, F, l), \; A(\vs, F, l) \big]$, where $\displaystyle \vs_{w_i\rightarrow 0}$ denotes the token $w_i$ in $\vs$ set to the zero embedding vector and $d$ denotes the attribution distance measure in Equation (\ref{eqn:ar}), described in the previous subsection. The tokens in $\vs$ are then sorted by descending values of $\displaystyle I_{\evw_i}$. Thus, we estimate words that are \textit{likely} to result in large attribution distances and prioritize those for substitutions towards building explanation attacks.\par%
\paragraph{Step 2: Candidate selection and substitution.} The second step of TEA substitutes each highest ranked token in $\vs$, computed in \textbf{Step 1}, with a token from a candidate set $\displaystyle \sC$, in descending importance order. Each highest ranked token has its separate candidate set $\displaystyle \sC$. TEA extracts these candidate sets by masking them and querying a transformer-based masked language model (MLM). In order to keep the computational costs low, we utilize the DistilBERT pretrained masked language model \citep{distilbert}, a BERT-MLM with significantly fewer parameters and more computationally efficient. Also, at most $\displaystyle n {}={} |\vs|  \cdot  \rho_{max}$ words are substituted.\par%
In order to further reduce computational cost, TEA uses batch masking. Thus, instead of masking each token separately and querying the MLM for candidates, the first $\displaystyle n_b {}={} |\vs| \cdot \rho_{b}$ most important tokens are masked at once and the language model is queried for candidates for all of these masked tokens. Here, $n_b$ denotes the number, $\rho_b$ the ratio of tokens in $\vs$ to be masked at once. For instance, during AR estimation of a 100 word text sample, given $\rho_{max}=0.15$ and $\rho_b=0.05$, the MLM is queried only $\displaystyle (100 \cdot 0.15)/(100 \cdot 0.05)=3$ times with batch masking instead of $100 \cdot 0.15 = 15$ times without it. We have compared the runtime of TEA using non-distilled \citep{bert} and distilled \citep{distilbert} BERT MLMs, with and without batch masking, and found considerable performance increase with batch masking and distillation. The results are reported in Section \ref{sec:experiments}.%
%
%
%
%

%% file: figures/algorithms.tex
\begin{algorithm}[t]
\caption{\texttt{TransformerExplanationAttack}}\label{alg:tea}
\textbf{Input}: Input sentence $\displaystyle \vs$ with predicted class $\displaystyle l$, classifier $\displaystyle F$, attribution $\displaystyle A$, attribution distance $\displaystyle d$, DistilBERT-MLM $L$, number of candidates $N$, maximum perturbation ratio $\displaystyle \rho_{max}$, batch masking ratio $\rho_b$\\
\textbf{Output}: Adversarial sentence $\displaystyle \vs_{\mathrm{adv}}$
\begin{algorithmic}[1]
\State $\displaystyle \vs_\mathrm{adv} \gets \vs$, $d_{max} \gets 0$, $r \gets 0$
\For{$\displaystyle w_i \in \vs$}
    \State $\displaystyle I_{w_i} {}={} d \big[ A(\vs_{w_i\rightarrow 0}, F, l), \; A(\vs, F, l) \big]$
\EndFor
\State{$\vs_B \gets \langle \vs_{1...b}, \vs_{b+1...2b}, ..., \vs_{|\vs|-b+1...|\vs|} \rangle$ with $I_{w_{b-1}} \geq I_{w_b} \; \forall j \in \{2,...,|\vs_B|\}$ and $\forall b \in \{1,..,|\vs_j|\}$}
\For{$\vs_b \in \vs_B$}
    \State $\mathbf{\sC_b} \gets L(\vs_{b \rightarrow [MASK]},  \vs_\mathrm{adv})$
    \For{$\displaystyle w_j \in \vs_b $}
        \If{$\displaystyle w_j \in \sS_{\mathtt{Stop\,words}}$}
            \State \textbf{continue}
        \EndIf
        \For{$\displaystyle \evc_k \in \sC_j$}
            \State \begin{small}{$\displaystyle \tilde{\vs}_{w_j\rightarrow\evc_k} \gets$ Replace $\displaystyle w_j$ in $\displaystyle \vs_\mathrm{adv}$ with $\displaystyle \evc_k$}\end{small}
            
            \If{$\displaystyle \argmax_{i \in \{1:|\sL|\}}F(\tilde{\vs}_{w_j\rightarrow\evc_k}) {}\neq{} l$}
            	\State \textbf{continue}
            \EndIf
            
            \State \begin{small}{$\tilde{d} = d \big[ A(\tilde{\vs}_{w_i\rightarrow \evc_k}, F, l),A(\vs, F, l) \big]$}\end{small}
        
            \If{$\displaystyle \tilde{d} > d_{max}$}
            
                \State $\displaystyle \vs_\mathrm{adv} \gets \tilde{\vs}_{w_i\rightarrow \evc_k}$
                \State $\displaystyle d_{max} \gets \tilde{d}$
                \State $\displaystyle r \gets r + 1$
            \EndIf
    
        \EndFor
        \If{$\rho {}={} \frac{r}{|\vs|} + 1 > \rho_{max}$}
            \State \textbf{break}
        \EndIf
    \EndFor
\EndFor
\end{algorithmic}
\end{algorithm}

%% file: src/results.tex
%
\input{figures/rho_plots.tex}%
\section{Experiments}%
\label{sec:experiments}%
In this section, we present our AR estimation experiments. Specifically, we describe the evaluation setup and results with our novel robustness definition. We show that TEA consistently outperforms our direct state-of-the-art competitor, \textsc{TextExplanationFooler} (TEF) in terms of the Lipschitz attribution robustness constant \textit{k} described in Section \ref{sec:methods}. Thus, we convey that TEA extracts smoother adversarial samples that are able to alter attributions more significantly than TEF. Finally, we compare the runtime of TEA to TEF and show that TEA achieves comparable runtimes, while still outperforming TEF in the previously mentioned aspects.\par%
\subsection{Setup}%
We evaluate the robustness constant \textit{k} estimated by TEA on the AG's News \citep{agnews_mr}, MR Movie Reviews \citep{agnews_mr}, IMDB Review \citep{imdb}, Yelp \citep{yelp} and the Fake News datasets \footnote{https://www.kaggle.com/c/fake-news/data}. We train a CNN, an LSTM, an LSTM with an attention layer (LSTMAtt), a finetuned BERT \citep{bert}, RoBERTA \citep{roberta} and XLNet \citep{xlnet} classifier for each dataset. A detailed description of these can be found in the appendix. We estimate the robustness of the Saliency (S), Integrated Gradients (IG) and Attention (A) attribution methods. The CNN and LSTM architectures are only used in combination with S and IG, the remaining LSTMAtt, BERT, RoBERTA and XLNet are used with all three attribution methods. Thus, we evaluate $2 \cdot 2 + 4 \cdot 3 = 16$ combinations of models and attributions for each dataset.\par%
We vary the $\rho_{max}$ parameter of TEA between $0.01$ and $0.4$. A value of $\rho_{max}$ does not necessarily lead to the actual perturbed ratio of tokens $\rho$ to be $\rho = \rho_{max}$, due to the prediction constraint. We set the batch masking size $\rho_b = \min(\rho_{max}, 0.15)$, as the MLM we use was trained by masking approximately 15\% of the tokens \citep{distilbert}. We set $|\sC| = 15$ for each run. Larger values do not tend to result in better estimation in terms of $k$, but to significantly higher attack runtimes. Moreover, this makes our experiments comparable to TEF \cite{ivankay2021fooling}.\par%
Our attack and experiments are implemented in PyTorch \citep{pytorch}, utilizing the Hugging Face Transformer library \citep{huggingface}, Captum \citep{captum} and SpaCy \citep{spacy}. We run each experiment on an NVIDIA A100 GPU with three different seeds and report the average results.\par%
\input{figures/relative_auc_plots.tex}%
\subsection{Results}
We report the following metrics as functions of the true perturbed ratio $\rho$. The average PCC values of original and adversarial attribution maps indicate the amount of change in explanations. Lower values correspond to larger attribution changes. The input distance between text samples is captured by the semantic textual similarity values of the original and adversarial samples, measured by the cosine similarity between the corresponding USE \citep{use} and MiniLM \citep{tse} sentence embeddings ($STS_{USE}$ and $STS_{MiniLM}$), as well as the relative perplexity increase ($\Delta_{PP}$). Moreover, the average increase in number of grammatical errors ($GE$) after perturbation is also reported. Using these values, we report the estimated Lipschitz robustness constants $k_{USE}$, $k_{MiniLM}$ and $k_{PP}$, according to Equation (\ref{eqn:expectedar}). In each of these, the scaled PCC is used as attribution distance. We compare these metrics for both our novel TEA algorithm and the direct competitor method TEF \citep{ivankay2021fooling}. Figure \ref{fig:detailedtightness} reports these metrics as a function of the true perturbed token ratio $\rho$. The continuous lines contain the reported metrics for our TEA attack, the dashed lines for the competitor TEF. The figure shows that TEA perturbations are able to alter explanations more (lower average PCC values), and they do so with adversarial samples equally or more semantically similar to the original inputs than TEF (higher average $STS$, lower average $PP$ and $GE$ values). Moreover, the perplexity increase is consistently lower for TEA perturbations, leading to more fluent adversarial samples. This is well-captured by resulting robustness constants $k$, which are consistently higher for TEA than TEF, showing both that our AR definition of Equation (\ref{eqn:ar}) is a suitable indicator for AR in text classifiers, and that TEA estimates this robustness better than the state-of-the-art TEF attack. The rest of the results is reported in the appendix.\par%
\input{figures/timing_plots.tex}%
In order to quantify the overall performance of TEA over the whole operation interval of $\rho$, we compute the area under the estimated $k - \rho$ curves ($2^{\mathrm{nd}}$ column in Figure \ref{fig:detailedtightness}). These are calculated as the integral $\mathrm{AUC}_k {}={} \int_{\rho} k(A, F) d\rho$. High $\mathrm{AUC}_k$ values correspond to high $k$-values, thus low overall robustness of attributions. We then compare the resulting $\mathrm{AUC}_k$ estimated with our TEA algorithm to the competitor method TEF. Figure \ref{fig:aucs} shows the relative increase of AUC when estimating with TEA rather than TEF, for each of the 16 combinations of models and attribution methods for a given dataset. For instance, a value of $0.5$ indicates a relative increase of 50\% in estimated $\mathrm{AUC}_k$, i.e. if TEF results in $\mathrm{AUC}_k=1.0$, TEA yields $\mathrm{AUC}_k=1.5$. We plot the $\mathrm{AUC}_k$ increase estimated with the semantic textual similarities from USE ($\mathrm{AUC}_k^{USE}$), MiniLM ($\mathrm{AUC}_k^{MiniLM}$) and with the relative perplexity increase ($\mathrm{AUC}_k^{PP}$) in the denominator of $k$. The attribution distance in the numerator of $k$ is set to the PCC described in Section (\ref{sec:methods}). We observe an increase in $\mathrm{AUC}_k$ of $0.3-0.5$ in case of USE and TSE, and $0.5-1.5$ in case of PP for most models, attribution maps and datasets. This further shows that TEA consistently yields higher robustness constants $k$ than TEF, providing better perturbations that alter attributions more while being more fluent and less perceptible.\par%
Querying transformer-based masked language models (MLMs) is computationally expensive. Naively substituting the synonym extraction from TEF with an MLM-based candidate extraction results in a significant increase in estimation time. Therefore, we use the methods described in Section \ref{sec:methods} to achieve comparable estimation time in our TEA algorithm and TEF. Figure \ref{fig:timings} contains the per-sample attack time for TEF (0), TEA with the non-distilled BERT MLM (1), TEA with DistilBERT MLM (2) and our final TEA algorithm with DistilBERT MLM and batch masking (3), for $\rho_{max} \in \{0.1, 0.25\}$. We observe that (1) results in a significant increase in mean estimation time by a factor of around 2 compared to (0) on both a smaller, medium and a larger datasets. Using (2) for estimating AR decreases the runtime by a large margin compared to (1). Finally, when applying both a distilled MLM and batch masking - TEA (3), the per-sample attack time is comparable to the baseline TEF, while maintaining better AR estimation performance.%
%

%% file: figures/rho_plots.tex
%
%
\begin{figure*}[t]
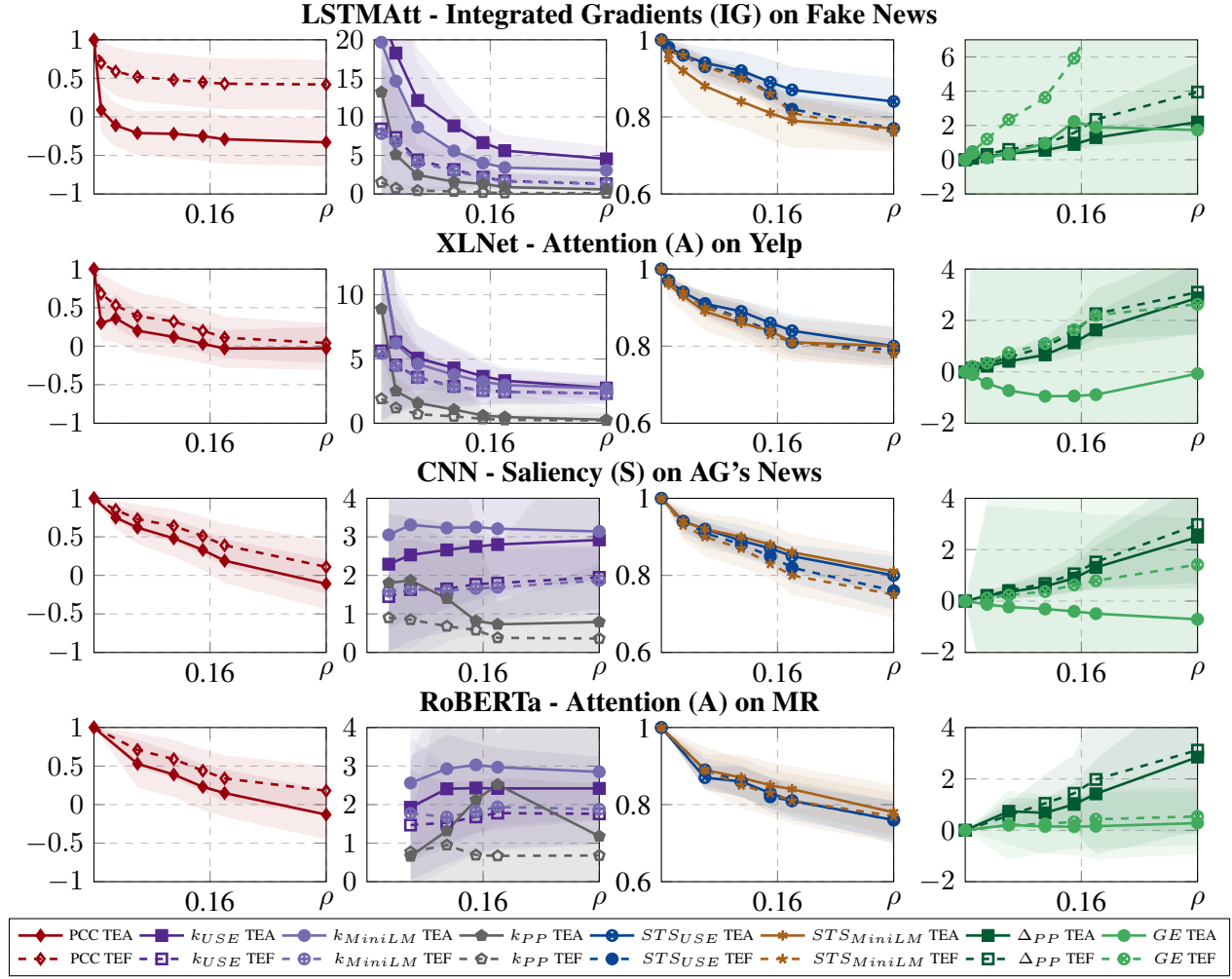
%
\setlength{\plotwidth}{0.295\textwidth}%
\setlength{\plotheight}{0.23\textwidth}%
\addtolength{\tabcolsep}{-12pt}  
\addtolength{\rowsepoffset}{-5pt} 
\centering%
    \begin{tabular}{cccc}%
    %
    %
    %
    \multicolumn{4}{c}{\textbf{LSTMAtt - Integrated Gradients (IG) on Fake News}}\\[\rowsepoffset]%
    \setval{colorpalette = r3}\setval{y_min = -1.0}\setval{y_max = 1.0}%
    \corrplot{fakenews/lstmatt/ig}
    & \setval{colorpalette = v3}\setval{y_min = 0.0}\setval{y_max = 20.0}%
    \kplot{fakenews/lstmatt/ig}
    & \setval{colorpalette = b3}\setval{y_min = 0.6}\setval{y_max = 1.0}%
    \stsplot{fakenews/lstmatt/ig}
    & \setval{colorpalette = g3}\setval{y_min = -2.0}\setval{y_max = 7.0}%
    \ppplot{fakenews/lstmatt/ig}%
    \\[-6pt]%
    %
    \multicolumn{4}{c}{\textbf{XLNet - Attention (A) on Yelp}}\\[\rowsepoffset]%
    \setval{colorpalette = r3}\setval{y_min = -1.0}\setval{y_max = 1.0}%
    \corrplot{yelp/xlnet/a}
    & \setval{colorpalette = v3}\setval{y_min = 0.0}\setval{y_max = 12.0}%
    \kplot{yelp/xlnet/a}
    & \setval{colorpalette = b3}\setval{y_min = 0.6}\setval{y_max = 1.0}%
    \stsplot{yelp/xlnet/a}
    & \setval{colorpalette = g3}\setval{y_min = -2.0}\setval{y_max = 4.0}%
    \ppplot{yelp/xlnet/a}%
    \\[-6pt]%
    %
    \multicolumn{4}{c}{\textbf{CNN - Saliency (S) on AG's News}}\\[\rowsepoffset]%
    \setval{colorpalette = r3}\setval{y_min = -1.0}\setval{y_max = 1.0}%
    \corrplot{agnews/cnn/s}
    & \setval{colorpalette = v3}\setval{y_min = 0.0}\setval{y_max = 4.0}%
    \kplot{agnews/cnn/s}
    & \setval{colorpalette = b3}\setval{y_min = 0.6}\setval{y_max = 1.0}%
    \stsplot{agnews/cnn/s}
    & \setval{colorpalette = g3}\setval{y_min = -2.0}\setval{y_max = 4.0}%
    \ppplot{agnews/cnn/s}%
    \\[-6pt]%
    \multicolumn{4}{c}{\textbf{RoBERTa - Attention (A) on MR}}\\[\rowsepoffset]%
    \setval{colorpalette = r3}\setval{y_min = -1.0}\setval{y_max = 1.0}%
    \corrplot{mr/roberta/a}
    & \setval{colorpalette = v3}\setval{y_min = 0.0}\setval{y_max = 4.0}%
    \kplot{mr/roberta/a}
    & \setval{colorpalette = b3}\setval{y_min = 0.6}\setval{y_max = 1.0}%
    \stsplot{mr/roberta/a}
    & \setval{colorpalette = g3}\setval{y_min = -2.0}\setval{y_max = 4.0}%
    \ppplot{mr/roberta/a}%
    \\[\rowsepoffset]%
    %
    \multicolumn{4}{c}{\allplotlegend}\\[\rowsepoffset]%
    \end{tabular}%
\caption{AR metrics as functions of the ratio of perturbed tokens $\rho$. We plot the mean and standard deviation of the Pearson correlations (PCC) between original and adversarial attributions, semantic similarities ($STS$), relative perplexity increase ($\Delta_{PP}$), increase of number of grammatical errors ($GE$) in original and adversarial text inputs and the estimated AR robustness constants ($k$). We compare these values for our novel \textsc{TransformerExplanationAttack} (TEA - continuous lines) and the current state-of-the-art \textsc{TextExplanationFooler} (TEF - dashed lines). We observe consistent improvement of robustness estimation with TEA compared to TEF, reflected in higher resulting $k$-values. This is attributed to both lower PCC values, higher semantic similarities of perturbed sentences to the original ones and lower adversarial perplexity of TEA perturbations.}%
\label{fig:detailedtightness}%
\addtolength{\tabcolsep}{12pt}  
\addtolength{\rowsepoffset}{5pt} 
\end{figure*}%

%% file: figures/relative_auc_plots.tex
\begin{figure*}[t]
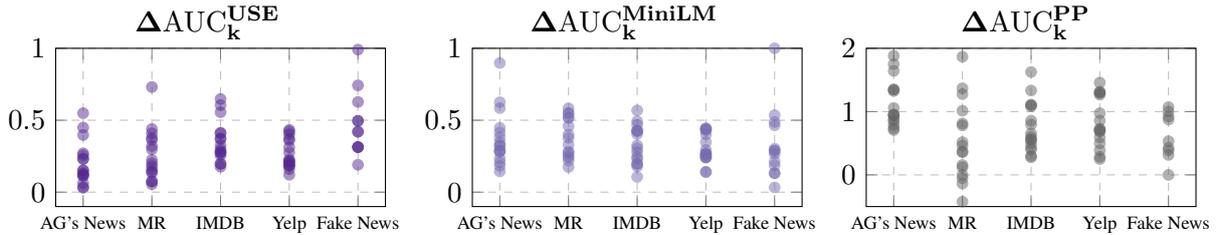
%
\setlength{\plotwidth}{0.37\textwidth}
\setlength{\plotheight}{0.23\textwidth}
\addtolength{\tabcolsep}{-9pt}  
\addtolength{\rowsepoffset}{-5pt} 
\centering%
    \begin{tabular}{ccc}%
    $\mathbf{\Delta{\mathrm{AUC}_k^{USE}}}$ & $\mathbf{\Delta{\mathrm{AUC}_k^{MiniLM}}}$ &$\mathbf{\Delta{\mathrm{AUC}_k^{PP}}}$\\[\rowsepoffset]%
    \setval{y_min = -0.1}\setval{y_max = 1.0}%
    \relativescatterplot{mlm_distil_fast}{et_use}{v30}
    & \setval{y_min = -0.1}\setval{y_max = 1.0}%
    \relativescatterplot{mlm_distil_fast}{et_tse}{v31}
    & \setval{y_min = -0.5}\setval{y_max = 2.0}%
    \relativescatterplot{mlm_distil_fast}{et_p}{v32}\\[\rowsepoffset]%
    \end{tabular}
\caption{Relative increase $\Delta$ of $\mathrm{AUC}_k$ when estimating the robustness constants $k$ (Equation \ref{eqn:expectedar}) with TEA compared to TEF. Each point corresponds to one of the 16 combinations of model and attribution method, on the indicated dataset. The $k$-values are estimated with the PCC as attribution similarity, varying the input distance measures $d_s$ as described in Section \ref{subsec:distances}. We observe a relative increase of $0.3-1.5$ for almost all models, attribution maps and datasets evaluated on. This shows that TEA consistently provides better perturbations that alter attributions more while being more fluent and semantically similar to the unperturbed input.}%
\label{fig:aucs}%
\addtolength{\tabcolsep}{9pt}  
\addtolength{\rowsepoffset}{5pt} 
\end{figure*}%

%% file: figures/timing_plots.tex
\begin{figure*}[t]
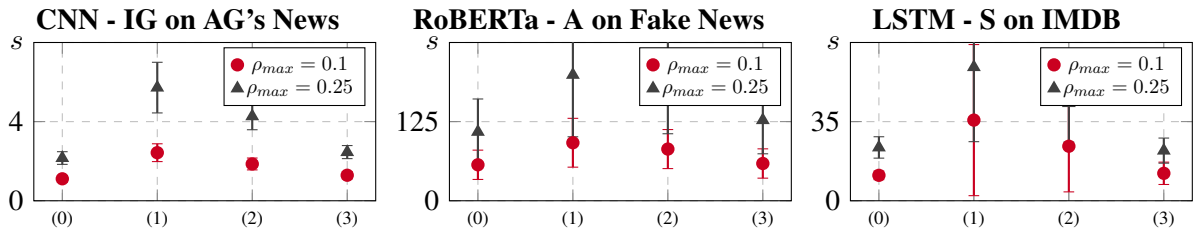
%
\setlength{\plotwidth}{0.38\textwidth}%
\setlength{\plotheight}{0.23\textwidth}%
\addtolength{\tabcolsep}{-7pt}  
\addtolength{\rowsepoffset}{-5pt} 
\centering%
        \begin{tabular}{ccc}%
        \textbf{CNN - IG on AG's News} & \textbf{RoBERTa - A on Fake News} & \textbf{LSTM - S on IMDB}\\[\rowsepoffset]%
        \setval{y_min = 0.0}\setval{y_max = 8.0}\setval{y_mid_tick = 4}%
        \timinglineplot{agnews}{cnn}{ig}%
        & \setval{y_min = 0.0}\setval{y_max = 250}\setval{y_mid_tick = 125}%
        \timinglineplot{fakenews}{roberta}{a}%
        & \setval{y_min = 0.0}\setval{y_max = 70.0}\setval{y_mid_tick = 35}%
        \timinglineplot{imdb}{lstm}{s}\\[\rowsepoffset]%
        \end{tabular}%
\caption{Per-sample runtime (s) of our AR estimator algorithm versions. TEA (3), with a distilled MLM and batch masking, achieves comparably fast estimation as TEF (0), while TEA with a non-distilled BERT MLM (1) is the slowest estimator, with a relative increase in runtime of approx. 1.5-2.5 compared to TEF. Distillation of the MLM (2) improves the runtime by around 25-35\% compared to (1).}%
\label{fig:timings}%
\addtolength{\tabcolsep}{7pt}  
\addtolength{\rowsepoffset}{5pt} 
\end{figure*}%

%% file: src/conclusion.tex
\section{Conclusion}%
\label{sec:conclusion}%
In this work, we introduced a novel definition of attribution robustness in text classifiers, derived from the notion of Lipschitz-continuity. Crucially, our definition incorporates the size of the perturbations, which contributes significantly to perceptibility. To this end, we introduce semantic textual similarity measures, the relative perplexity increase and the number of grammatical errors as ways to effectively quantify perturbation size in text. Moreover, we introduced \textsc{TransformerExplanationAttack}, a novel state-of-the-art attack method that results in a tighter estimator for attribution robustness in text classification problems. It is a black box estimator that utilizes a distilled MLM with batch masking to extract good adversarial perturbations with small computational overhead. Finally, we showed that TEA outperforms current attack methods by altering explanations more significantly using less perceptible perturbations.\par%
In future work, we plan to examine the robustness of a wider variety of attributions, develop methods that improve explanation robustness and adapt the novel techniques to real-life scenarios.%